\documentclass[10pt,journal,compsoc]{IEEEtran}
\usepackage{amsmath,amsfonts}
\usepackage{algorithmic}
\usepackage{algorithm}
\usepackage{array}
\usepackage{bbm}
\usepackage{dsfont}

\usepackage[caption=false,font=footnotesize,labelfont=rm,textfont=rm]{subfig}
\usepackage{textcomp}
\usepackage{stfloats}
\usepackage{url}
\usepackage{verbatim}
\usepackage{graphicx}

\usepackage{xcolor}
\usepackage{xpatch}
\makeatletter
\def\changeBibColor#1{
\in@{#1}{bonawitz2017practical, xu2023edge, yao2024wireless, dwork2006calibrating, dwork2014algorithmic}
\ifin@\color{black}\else\normalcolor\fi
}
\xpatchcmd\@bibitem
{\item}
{\changeBibColor{#1}\item}
{}{\fail}
\xpatchcmd\@bibitem
{\item}
{\changeBibColor{#1}\item}
{}{\fail}
\makeatother

\usepackage{bm}
\usepackage{multirow}
\usepackage{makecell}

\usepackage{caption}
\usepackage{amsfonts,amssymb}
\usepackage{graphicx}
\usepackage{epstopdf}
\usepackage{mathrsfs}
\usepackage{amsmath}
\usepackage{latexsym}
\usepackage{booktabs}
\usepackage{float}
\usepackage{algorithm, algorithmic}
\usepackage{bbm}
\usepackage[amsmath,thmmarks]{ntheorem}
\usepackage{bm}
\usepackage{setspace}

\definecolor{blue}{named}{black}

\ifCLASSOPTIONcompsoc
  \usepackage[nocompress]{cite}
\else
  \usepackage{cite}
\fi
\ifCLASSINFOpdf
\else
\fi
\begin{document}
\title{Optimizing Communication and Device Clustering for Clustered Federated Learning with Differential Privacy  \vspace*{-0em}}

\author{{{Dongyu Wei}, {Xiaoren Xu},  {Shiwen Mao,} \emph{Fellow,~IEEE},  {Mingzhe Chen,} \emph{Senior Member,~IEEE}}% <-this % stops a space
\IEEEcompsocitemizethanks{\IEEEcompsocthanksitem D. Wei, X. Xu and M. Chen are with the Department of Electrical and Computer Engineering, University of Miami, Coral Gables, FL, 33146, USA. \protect E-mail: \protect\url{dongyu.wei@miami.edu}; \protect\url{xiaoren.xu@miami.edu};
\protect\url{mingzhe.chen@miami.edu}

\IEEEcompsocthanksitem Shiwen Mao is with the Department of Electrical and Computer Engineering, Auburn University, Auburn, AL, 36849, USA. \protect E-mail: \protect\url{smao@ieee.org}

\IEEEcompsocthanksitem Mingzhe Chen is also with the Frost Institute for Data Science and Computing, University of Miami, Coral Gables, FL 33146, USA.}% <-this % stops a space
\thanks{}}

\markboth{IEEE TRANSACTIONS ON MOBILE COMPUTING}%
{Shell \MakeLowercase{\textit{et al.}}: Bare Advanced Demo of IEEEtran.cls for IEEE Computer Society Journals}
%\vspace{-0.5 cm}


%\pagestyle{empty}  % no page number for the second and the later pages
\thispagestyle{empty} % no page number for the first page

%\vspace{-0.5 cm}
\IEEEtitleabstractindextext{
\begin{abstract}
In this paper, a secure and communication-efficient clustered federated learning (CFL) design is proposed. In our model, several base stations (BSs) with heterogeneous task-handling capabilities and multiple users with non-independent
and identically distributed (non-IID) data jointly perform CFL training incorporating differential privacy (DP) techniques. Since each BS can process only a subset of the learning tasks and has limited wireless resource blocks (RBs) to allocate to users for federated learning (FL) model parameter transmission, it is necessary to jointly optimize RB allocation and user scheduling for CFL performance optimization. Meanwhile, our considered CFL method requires devices to use their limited data and FL model information to determine their task identities, which may introduce additional communication overhead. We formulate an optimization problem whose goal is to minimize the training loss of all learning tasks while considering device clustering, RB allocation, DP noise, and FL model transmission delay.
%and erroneous device clustering.   
%we present a comprehensive framework for implementing Clustered Federated Learning (CFL) over wireless networks, addressing key challenges in multi-task clustering, resource allocation, and security. Unlike existing approaches, our model considers the heterogeneous task-handling capabilities of base stations (BSs) and integrates differential privacy (DP) to protect sensitive information during model transmission. 
%We propose a joint optimization problem to minimize the CFL training loss while ensuring transmission efficiency and privacy preservation. 
To solve the problem, 
%we derive a closed-form expression for the expected convergence rate, linking transmission requirements and DP variance to CFL performance.
%The problem is then addressed using 
we propose 
a novel dynamic penalty function assisted value decomposed multi-agent reinforcement learning (DPVD-MARL) algorithm that enables distributed BSs to independently determine their connected users, RBs, and DP noise of the connected users but jointly minimize the training loss of all learning tasks across all BSs. Different from the existing MARL methods that assign a large penalty for invalid actions, we propose a novel penalty assignment scheme that assigns penalty depending on the number of devices that cannot meet communication constraints (e.g., delay), which can guide the MARL scheme to quickly find valid actions, thus improving the convergence speed. Simulation results show that the DPVD-MARL can improve the convergence rate by up to 20\% and the ultimate 
accumulated rewards by 15\% compared to independent Q-learning.
\end{abstract}

}

\theoremstyle{definition}
\newtheorem{theorem}{Theorem}
\newtheorem{proposition}{Proposition}
\newtheorem{lemma}{Lemma}
\newtheorem{proof}{Proof}

\maketitle

\section{Introduction}\label{Introduction}

Federated learning (FL) enables devices to collaboratively train a machine learning model while keeping data at edge devices, thus improving device data privacy and reducing communication costs \cite{kang2020reliable, chen2021communication, xu2021learning}. 
However, standard FL may not be effective for learning tasks where devices have non-independent and identically distributed (non-IID) data \cite{lu2024data, itahara2021distillation}.
To address this issue, clustered federated learning (CFL) clusters devices into several groups and each group consists of  the devices with a similar data distribution. 
Hence, CFL allows edge devices with similar data distributions to collaboratively train a machine learning model so as to address the non-IID issues.  
However, deploying CFL over wireless networks faces several challenges. 
First, since devices do not share data with the server, the server has limited data information to determine cluster identities of devices, thus introducing erroneous device clustering and reducing learning performance.
Second, the implementation of CFL requires the server to update the models of several device groups. 
Due to limited energy and wireless resources, each server may not be able to process all the model updates from many groups. 
Hence, it is necessary to develop efficient resource management and scheduling strategies to improve the FL model and update efficiency. 
Finally, CFL introduces new privacy and security vulnerabilities since an attacker can attack not only model parameters but also clustering related information.  

Several existing works \cite{albaseer2021client, xu2024clustered, taik2022clustered, xia2022optimization, gong2022adaptive, he2023stackelberg} have studied the optimization of wireless resource management and user scheduling for CFL. In particular, the authors in \cite{albaseer2021client} designed a distributed device selection scheme for CFL in order to reduce training time.
In \cite{xu2024clustered}, the authors jointly optimized resource allocation and device clustering using convex optimization and coalition formation games in order to improve bandwidth efficiency and training performance. The work in \cite{taik2022clustered} proposed a CFL framework, consisting of a cluster-head selection mechanism and resource block allocation scheme to improve communication efficiency and learning accuracy under non-IID data. Similarly, the authors in \cite{xia2022optimization} also developed a joint optimization framework for clustering and resource allocation, using coalition formation and convex optimization to efficiently cluster devices with similar data distributions and optimize communication performance. The authors in \cite{gong2022adaptive} presented a CFL framework that reduces communication overhead by dynamically grouping clients based on partial model weights. In \cite{he2023stackelberg}, the authors proposed a CFL framework that leverages gradient similarity for device clustering and a three-stage Stackelberg game model for resource allocation.
\textcolor{black}{However, most of these existing approaches \cite{xu2024clustered, taik2022clustered, xia2022optimization, gong2022adaptive, he2023stackelberg} were focused on the CFL frameworks where device clustering is implemented by a centralized base station (BS) or parameter server, thus requiring devices to transmit device clustering related information (e.g., local model parameters) to the BS or the server, resulting in high communication and computing overhead. Meanwhile, all these approaches \cite{albaseer2021client, xu2024clustered, taik2022clustered, xia2022optimization, gong2022adaptive, he2023stackelberg} assumed that each server can process all the learning tasks, which is impractical in real-world applications since different servers may have different task processing priorities, computing power, and wireless resources. In addition, most of these works \cite{xu2024clustered, taik2022clustered, xia2022optimization, gong2022adaptive} only considered FL model parameter transmission from users to servers but ignored the model parameter transmission from servers to users. Hence, these methods \cite{xu2024clustered, taik2022clustered, xia2022optimization, gong2022adaptive} may not be suited for CFL since a CFL server needs to transmit multiple FL models and the transmission delay cannot be ignored. Finally, none of these works \cite{albaseer2021client, xu2024clustered, taik2022clustered, xia2022optimization, gong2022adaptive, he2023stackelberg} considered the data privacy issues of CFL. Compared to standard FL, the devices in CFL must determine their cluster identities, thus may have more sensitive information to be protected. }

Recently, several existing works such as \cite{he2023stackelberg, zhou2023reinforcement, hafid2024centralized, sun2022fair, gauthier2023networked, yang2024privacy} have studied the use of reinforcement learning (RL) in FL to address various challenges, including client selection, resource management, and performance optimization. In particular, the authors in \cite{he2023stackelberg} designed a hierarchical RL algorithm to optimize resource allocation and improve training efficiency in CFL for heterogeneous UAV swarms. The work in \cite{zhou2023reinforcement} proposed a deep reinforcement learning-based client selection framework for CFL to minimize job completion time by optimizing intra-cluster and inter-cluster operations. In \cite{hafid2024centralized}, the authors integrated deep RL with hybrid FL to enhance energy efficiency and fault tolerance in autonomous swarm robotics. In \cite{sun2022fair}, the authors introduced a policy gradient-based RL framework, to adjust FL aggregation weights across clients to enhance fairness and convergence. The work in \cite{gauthier2023networked} presented an RL-based personalized FL framework that adapts inter-cluster learning parameters to improve local model performance and overall training accuracy. The authors in \cite{yang2024privacy} developed a graph neural network (GNN)-integrated proximal policy optimization (PPO) algorithm to enhance privacy preservation and robust aggregation in collaborative FL while reducing computational overhead and maintaining model accuracy.
\textcolor{black}{However, most of these works \cite{he2023stackelberg, zhou2023reinforcement, hafid2024centralized} were focused on optimizing single decision variables such as resource management or user association, and they may not be effective for the joint optimization of multiple decision variables since increasing the number of variables will significantly increases the action and state space of the RL algorithm, thus resulting in slow RL convergence and high training complexity. Meanwhile, the works \cite{hafid2024centralized, sun2022fair, gauthier2023networked} did not consider the collaboration among RL agents which prevents these algorithms from finding globally optimal solutions, and may introduce potential conflicts or interferences, thus undermining the CFL efficiency. Finally, all of these works~\cite{he2023stackelberg, zhou2023reinforcement, hafid2024centralized, sun2022fair, gauthier2023networked, yang2024privacy} employed standard RL algorithms for performance optimization of CFL and hence, they did not consider how FL model training and transmission constraints (e.g., delay) affect the RL performance.}

\textcolor{black}{The aim of this work is to design a novel communication-efficient, secure, and multi-task CFL framework that overcomes the aforementioned limitations. The main contributions include:
\begin{itemize}
    \item Unlike conventional CFL frameworks \cite{ghosh2022efficient} that only consider one parameter server determining the task identities, we propose a novel CFL framework in which edge users determine their task identities, and transmit their locally trained models with differential privacy (DP) to BSs. Each BS, handling different learning tasks, aggregates local models from users to obtain global models and transmits them back to the users. Due to limited wireless resources, each BS must select appropriate users to join FL training and allocate them appropriate  RBs for FL parameter transmission.   
    This task is formulated as an optimization problem aiming to minimize the weighted sum of CFL training loss and data privacy while meeting transmission and DP security constraints. 
    \item To solve this problem, we first perform a fundamental analysis on the expected convergence rate of our proposed CFL algorithms. However, we cannot directly use the analytical result to jointly optimize user association, RB allocation, and DP noise, since the convergence analysis contains unknown parameters such as clustering error rates in CFL. 
    To this end, we propose a novel dynamic penalty function assisted value decomposition and optimization based multi-agent reinforcement learning (DPVD-MARL) algorithm.
    The proposed algorithm allows each BS to independently manage user association, allocate RBs, and adjust DP noise variance levels, while collaboratively optimizing the overall training loss across all learning tasks and BSs. 
    Specifically, in the proposed DPVD-MARL method, each action only determines user associations. The RB allocation and the DP noise are determined by the optimization methods. This way, the action space can be significantly reduced, thus improving RL training speed.  
    \item To further improve RL convergence speed, we propose a progressive invalid action penalty scheme based on the number of users violating the communication constraints (e.g., delay). Our proposed method is different from existing MARL approaches \cite{lee2022radio, hazarika2024quantum} that imposed a fixed, large penalty for any invalid actions, which will discourage exploration and may lead to suboptimal policies.
\end{itemize}}
\noindent Simulation results show that the DPVD-MARL can improve the convergence rate by up to 35\% and the ultimate accumulated rewards by 27\% compared to independent Q-learning.

The rest of the paper is organized as follows. The proposed CFL algorithm and problem formulation are detailed in Section II. The CFL convergence analysis is introduced in Section III. The DPVD-MARL based solution is discussed in Section IV. Simulation settings and results are presented in Section V. Conclusions are drawn in Section VI.

\section{System Model and Problem Formulation}\label{Proposed_Clustered_FL_System}

\begin{figure}[tp]
    \centering
\includegraphics[width=.48\textwidth]{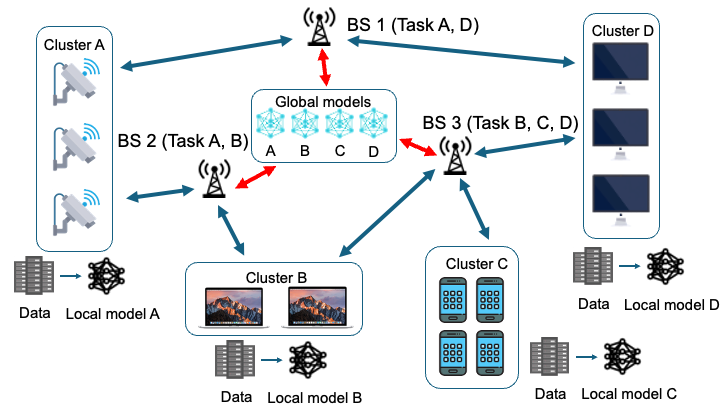}
    \caption{The architecture of the considered CFL framework.}
    \label{fig:system}
    \vspace{-0.3cm}
\end{figure}

\textcolor{black}{Consider a multi-task federated learning system where a set $\mathcal{B}$ of $B$ trusted BSs and a set $\mathcal{U}$ of $U$ users holding non-IID data cooperatively execute CFL algorithms, as shown in Fig.~\ref{fig:system}.}
% During transmission, we assume that an orthogonal frequency division multiple access (OFDMA) technique is used, in which each user occupies one resource block (RB).
% We also consider an eavesdropping probability vector $\mathbf{E_i}$ to represent the data leaking probability between BS $i$ and each user.
% In our model, users have different datasets, making the data distribution non-IID.
%Let $K$ be the total number of data distributions across all users, which indicates that $K$ learning tasks must be completed by the BSs and users. 
\textcolor{black}{We assume that the BSs and users must complete a set $\mathcal{K}$ of $K$ learning tasks and the total number of data distributions of users is $K$.}
Each BS~$b$ can perform a subset $\mathcal{K}_{b}$ of $K_{b}$ learning tasks such that BS~$b$ will only select the users that have datasets for these tasks.  
Meanwhile, the learning tasks that two BSs execute may overlap, e.g., $\mathcal{K}_a = \{1, 2, 3\}$, $\mathcal{K}_b = \{1, 3, 5\}$, and hence $\mathcal{K}_a \cap \mathcal{K}_b = \{1, 3\}$. \textcolor{black}{Table \ref{tb:notations} provides a summary of the notations used hereinafter.}
% Consider $\mathcal{D}$ as the differential privacy (DP) matrix used for additive noise during data transmission.

\begin{table*}[tp]
\captionsetup{labelfont={color=black},textfont={color=black}}
\caption{List of Notations}
\label{tb:notations}
\centering
\fontsize{9}{9}\selectfont{
{\color{black}\begin{tabular}{|c||c|c||c|}
\hline
\textbf{Notation}         & \textbf{Description}  & \textbf{Notation}         & \textbf{Description}  \\ \hline
$B$ & Number of BSs & $\epsilon_i\left(t\right)$ & The task identity of user $i$ at iteration $t$ \\ \hline
$U$ & Number of users & $\alpha$ & CFL learning rate \\ \hline
$K$ & Total number of learning tasks & $a_{i,b}\left(t\right) \in \{0,1\}$ & Connection index between user $i$ and BS $b$ \\ \hline
$K_b$ & Number of learning tasks for BS $b$ & $R$ & Number of RBs \\ \hline
$\boldsymbol{w}_{k}^G\left(t\right)$ & The global model of task $k$ & $\boldsymbol{w}^i_{k}\left(t\right)$ & The local model of task $k$ at user $i$\\ \hline
$X_i$ & Number of data samples at user $i$ & $B^{\textrm{U}}$ & Uplink bandwidth \\ \hline
$\boldsymbol{n}_{i}\left(t\right)$ & DP noise applied by user $i$ & $I^n_{i,b}\left(t\right)$ & Inter-user inference at user $i$ \\ \hline
$c_{i,b,t}^\textrm{U}\left(\boldsymbol{r}_{i,b}\left(t\right)\right)$ & Uplink rate from user $i$ to BS $b$ & $r_{i,b}^n\left(t\right) \in \{0,1\}$ & RB allocation vector \\ \hline
$\gamma_D$ & Delay requirement & $\gamma_T$ & Time requirement \\ \hline
$\boldsymbol{w}^*_k$ & Optimal global FL model of task $k$ & $P_i$ & Transmit power of user $i$ 
\\ \hline
$h_{i,b}$ & Channel gain from user $i$ to BS $b$ & $W_0$ & Noise power spectral density \\ \hline
$d$ & The dimension of DP noise & $F\left(\boldsymbol{w}_{k}^G\left(t\right)\right)$ & CFL loss of global model of task $k$ \\ \hline
$\boldsymbol{s}_b(t) \in \mathbb{R}^{1 \times 2U}$ & State of each BS $b$ & $\boldsymbol{a}_b(t) \in \mathbb{R}^{1 \times U}$ & Action of each BS $b$ \\ \hline
$R\left(\boldsymbol{a}\left(t\right)|\boldsymbol{s}\left(t\right)\right)$ & Shared reward for all the BSs & $Q_{tot}\left(\boldsymbol{s}\left(t\right), \boldsymbol{a}\left(t\right)\right)$ & Global Q function \\ \hline
\end{tabular}}
}
\end{table*}

The BSs do not know the task identity of each device.
Each device must use its limited FL parameter information to determine its task identity and use a DP scheme to add artificial noise to its FL model parameters so as to improve the privacy of its transmitted FL model parameters.
Each BS must select its associated users to join the FL model training according to their task identities and allocate appropriate RBs to their associated users for FL model parameter transmission.
% To address the data heterogeneity problem, users can be divided into $K$ clusters based on the characteristics of their datasets.
% The devices with similar data distributions are clustered into a group and jointly perform FL training on each BS.
% In the model, we consider a general scenario where each device does not know the data distribution of other devices, and each BS also does not know the others' RB assignment strategies.
In this section, we first introduce our proposed clustered FL algorithms.
Then, we present the FL parameter transmission and DP models.
Finally, we will formulate our optimization problem. 

\subsection{General Procedure of Clustered FL Algorithms}

Since each BS $b$ implements $K_b$ learning tasks, it must implement $K_b$ clustered FL algorithms. 
% Denote $X_i = [x_{i1}, ..., x_{iK_i}]$ as the input data collected by user $i$ and let $y_i = [y_{i1}, ..., y_{iK_i}]$ be the corresponding output vector.
The training process of these clustered FL algorithms at BS $b$ is summarized as:% follows:
% We also assume that each BS has access to the same RB set $\mathcal{R}$.

\emph{1) BS $b$ initialization:} 
$K_b$ FL models are randomly initialized and broadcast to all users. 
% According to the model we introduced above, more than one BS may have the same cluster identity. 
Let $\boldsymbol{w}_{k}^G\left(t\right)$ be the globel FL parameters of task $k$ of BS $b$ at iteration $t$. 
The set of users belonging to task $k$ may vary according to the task identities determined by the users per iteration.

% \subsubsection{Parameter Aggregation and Integration}
% Each user receives FL model parameters from all BSs. 
% As the user may receive model parameters belonging to one cluster identity from more than one BS, we take an average on them to get aggregated FL parameter results of each cluster for users. 
% Assume that $w'_{k,i}\left(t\right)$ is the averaged parameters of cluster $k$ at user $i$. 
% Hence, $w'_{k,i}\left(t\right)$ can be given by: 
% \begin{equation}
%     w'_{k,i}\left(t\right) = \frac{1}{N_k}\sum_{1 \leq b \leq B}^{\exists k \in \mathcal{T}_b}{w_b(k, t)}, \label{eq:weight_avg}
% \end{equation}
% where $N_k$ is the number of BSs having cluster $k$ FL model parameters.

\emph{2) Determination of task identity:}
Since each user will receive the FL model parameters from all BSs and different BSs broadcast the same global FL model for a given task, it will receive $K$ FL models, i.e., $\left\{\boldsymbol{w}_{1}^G\left(t\right), \boldsymbol{w}_{2}^G\left(t\right), \ldots, \boldsymbol{w}_{K}^G\left(t\right)\right\}$. 
At the beginning, each user randomly initializes its task identity, as no prior information is available. Then, given these FL models, each user must determine its task identity and then update its corresponding FL model.  
The specific steps are summarized as follows:

\textbf{Step 1: Loss Function and Gradient Calculation:}
Let $\mathcal{X}_i$ be a set of $X_i$ data samples of user $i$. Each data sample $j$ of user $i$ consists of an input vector $\boldsymbol{x}_{i,j}$ and an output label $y_{i,j}$. 
User $i$ uses a mini-batch $\mathcal{X}_{i}\left(t\right)$ of $X_{i}\left(t\right)$ data samples to train each FL model $\boldsymbol{w}_{k}\left(t\right)$ and obtain the value of the loss function as follows:
\begin{equation}\label{eq:training_loss}
\mathcal{L}_{i,k}^{t}\left(\mathcal{X}_{i}\left(t\right)\right)=\!\!\!\!\sum_{\boldsymbol{x}_{i,j}\left(t\right)\in \mathcal{X}_{i}\left(t\right)} \!\!\!\!l\left(\boldsymbol{w}_k^G\left(t\right), \boldsymbol{x}_{i,j}\left(t\right), y_{i,j}\left(t\right)\right), \forall k \in \mathcal{K}.
\end{equation}
where $l\left(\boldsymbol{w}_k^G\left(t\right), \boldsymbol{x}_{i,j}\left(t\right), y_{i,j}\left(t\right)\right)$ is the loss value of model $\boldsymbol{w}_k^G\left(t\right)$ with respect to $\boldsymbol{x}_{i,j}\left(t\right)$ and $y_{i,j}\left(t\right)$.

Given the loss function of each task $k$, user~$i$ calculates the gradient $\nabla\mathcal{L}_{i,k}^{t}\left(\mathcal{X}_{i}\left(t\right)\right), \forall k \in \mathcal{K}$ of $\mathcal{L}_{i,k}^{t}\left(\mathcal{X}_{i}\left(t\right)\right)$ with respect to the global model $\boldsymbol{w}_k\left(t\right)$.

\textcolor{black}{\textbf{Step 2: Similarity Calculation:}
Different from the prior works \cite{sattler2020byzantine, mcmahan2017communication} that use only loss values resulting from different FL models to determine the task identity of users, here, we consider not only loss values but also the gradient descent direction. To calculate the gradient descent  direction, we first define $\Delta \boldsymbol{w}_{k}\left(t\right)$ as the gap between the global FL model $\boldsymbol{w}_k^G\left(t\right)$ and $\boldsymbol{w}_k^G\left(t-1\right)$ as
\begin{equation}\label{eq:gap}
    \Delta \boldsymbol{w}_{k}\left(t\right) = \boldsymbol{w}_k^G\left(t\right) - \boldsymbol{w}_k^G\left(t-1\right), \forall k \in \mathcal{K}.
\end{equation} 
We then calculate the similarity between the global FL model gradient direction $\nabla\mathcal{L}_{i,k}^{t}\left(\mathcal{X}_{i}\left(t\right)\right)$ and the local FL model gradient direction $\Delta \boldsymbol{w}_{k}\left(t\right)$ as follows:
\begin{equation}\label{eq:similarity}
    S_{i, k}\left(t\right)=\frac{\nabla\mathcal{L}_{i,k}^{t}(\mathcal{X}_{i}\left(t\right)) \cdot \Delta w_{k}\left(t\right)}{\left\vert\nabla\mathcal{L}_{i,k}^{t}(\mathcal{X}_{i}\left(t\right))\right\vert \times \left\vert\Delta w_{k}\left(t\right)\right\vert}, \forall k \in \mathcal{K}.
\end{equation}}

\textcolor{black}{\textbf{Step 3: Task Identity Determination:}
Let $\epsilon_i\left(t\right)$ be the task identity of user $i$ at iteration $t$. At the initialization stage (i.e., t=0), 
$\epsilon_i(0)$ is a random value. However, we must randomly select $K$ users and allocate one user to each learning task in order to ensure that each task is implemented by at least one user.
When $t > 0$, we have
\begin{equation} \label{eq:identity}
    \epsilon_i\left(t\right) = \underset{k \in \mathcal{K}}{\mathrm{argmax}} \left(\lambda S_{i, k}\left(t\right) + \left(1-\lambda\right) \left(-\mathcal{L}_{i, k}^{t}\left(\mathcal{X}_{i}\left(t\right)\right)\right)\right),
\end{equation}
where $\lambda$ is a parameter that controls the weights of $S_{i, k}\left(t\right)$ and $\mathcal{L}_{i, k}^{t}\left(\mathcal{X}_{i}\left(t\right)\right)$ for task identification.}

\emph{3) Local model update and upload:}
To ensure the privacy of FL model parameter transmission, each user $i$ will introduce DP noise to the gradient of its loss function \cite{wei2020federated}.
Let $\boldsymbol{n}_{i}\left(t\right) \sim N\left(0,\sigma^2_{i}\mathbf{I}_d\right)$ be the DP noise that is applied by user $i$, with $\sigma_i$ being the DP noise standard deviation of user $i$, and $d$ being the dimension of $\boldsymbol{n}_i\left(t\right)$. Then we have 
\begin{equation} \label{eq:updated_grad}
    \mathfrak{M}_{i,\epsilon_i\left(t\right)}^t\left(\mathcal{X}_i\right) = \nabla \mathcal{L}_{i, \epsilon_i\left(t\right)}^{t}\left(\mathcal{X}_i\right) + \boldsymbol{n}_{i}\left(t\right),
\end{equation}
where $\mathfrak{M}_{i,\epsilon_i\left(t\right)}^t\left(\mathcal{X}_i\right)$ is user $i$'s gradient with DP noise.
Assuming that $\left\|\nabla \mathcal{L}_{i, \epsilon_i\left(t\right)}^{t}\left(\mathcal{X}_i\right)\right\|_2 \leq M$, the privacy leakage of $\mathfrak{M}_{i,\epsilon_i\left(t\right)}^t\left(\mathcal{X}_i\right)$ defined under $\rho$-zCDP \cite{tavangaran2024differential} in every iteration is $\rho_i = 2\left(\frac{M}{X_i\sigma_{i}}\right)^2$.
% \begin{equation}\label{eq:rho}
%     \rho_i = 2(\frac{L}{X_i\sigma_{i}})^2.
% \end{equation}
Next, the local model $\boldsymbol{w}^i_{\epsilon_i\left(t\right)}\left(t+1\right)$ of user $i$ is updated by
\begin{equation} \label{eq:update}
    \boldsymbol{w}^i_{\epsilon_i\left(t\right)}\left(t+1\right) = \boldsymbol{w}^G_{\epsilon_i\left(t-1\right)}\left(t\right) - \alpha \mathfrak{M}_{i,\epsilon_i\left(t\right)}^t\left(\mathcal{X}_i\right),
\end{equation}
where $\alpha$ is the learning rate.
Then, user $i$ first broadcasts $\epsilon_i\left(t\right)$ to all BSs. Each BS determines its associated users based on the task identities, user locations, and wireless channel conditions. Afterwards, each user transmits its updated local model $\boldsymbol{w}^i_{\epsilon_i\left(t\right)}\left(t+1\right)$ to its connected BS. 
\textcolor{black}{Here, since the BSs are trusted, they will directly receive task identities of the users. If several BSs are untrusted, they may infer user data information from task identities of the users. To address this issue, we can use encryption methods \cite{bonawitz2017practical} to encrypt the task identities of the users such that the BSs use the encrypted task identities to aggregate models without knowing the real task identity information of the users.}
% \textcolor{red}{To reduce the transmission latency, analog transmission techniques such as over-the-air computation (AirComp) \cite{yao2024wireless} can further improve the transmission efficiency.}}

\emph{4) FL model aggregation:}
Let $a_{i,b}\left(t\right) \in \{0,1\}$ be a connection index between user $i$ and BS $b$, with $a_{i,b}\left(t\right) = 1$ indicating that user $i$ connects to BS $b$, and $a_{i,b}\left(t\right) = 0$, otherwise.
The model aggregation of task $k$ at BS $b$ is represented as
\begin{equation} \label{eq:updated_global}
    \boldsymbol{w}_{k,b}\left(t+1\right) = \frac{1}{N_{k,b}} \sum\limits_{i \in \mathcal{U}_{k,t}}a_{i,b}\left(t\right)X_i\boldsymbol{w}_k^i\left(t+1\right), \forall b \in \mathcal{B},
\end{equation}
where $\sum_{i=1}^Ua_{i,b}\left(t\right) \leq 1$, $\mathcal{U}_{k,t}$ is the set of users whose task identity is $k$, and $N_{k,b} = \sum\limits_{i \in \mathcal{U}_{k,t}}a_{i,b}\left(t\right)X_i$ is the total number of data samples that the users of BS $b$ use for model update of task $k$.
After aggregating the local FL models of the associated users, each BS will also share its FL models with other BSs so as to generate a common FL model for each task across all BSs. Hence, the global model aggregated by all the BSs is
\begin{equation}\label{eq:get_global}
    \boldsymbol{w}^G_k\left(t+1\right) = \frac{1}{\left|\mathcal{B}_k\right|\sum_{b \in \mathcal{B}_k}\!N_{k,b}}\!\sum\limits_{b \in \mathcal{B}_k}N_{k,b}\boldsymbol{w}_{k,b}\left(t+1\right), \forall k \in \mathcal{T},
\end{equation}
where $\mathcal{B}_k$ is the set of BSs that must execute task $k$.

Steps 2)-4) are repeated until the optimal FL models are found.
The procedure of our proposed clustered FL algorithm is summarized in Algorithm \ref{algo:clustering_federated_learning}.

\addtolength{\topmargin}{0.05in}
\setlength{\textfloatsep}{4pt}
\begin{algorithm}[t]
    \small
    \caption{Proposed Clustered Federated Learning}
    \begin{algorithmic}[1]\label{algo:clustering_federated_learning}
        \STATE \textbf{Input: } set of BSs $\mathcal{B}$, set of users $\mathcal{U}$, number of tasks $K$, number of clustering iterations $T$, set of task identities $\mathcal{K}$, initial global FL models of each BS $\left\{\boldsymbol{w}_{k}^G\left(0\right)| b \in \mathcal{B}, k \in \mathcal{K}_b\right\}$;
        \STATE \textbf{Initialization}: Each BS $b \in \mathcal{B}$ initializes $K_b$ FL models and each user $i \in \mathcal{U}$ initializes its task identity randomly;

        \FOR{$t = 1, 2, \hdots, T$}
            \STATE \underline{Each BS $b$:} broadcast $\left\{\boldsymbol{w}_{k}^G\left(t\right)|k \in \mathcal{K}_b\right\}$ to all devices;
            
            \FOR{\underline{Each user $i\in \mathcal{U}$} in parallel}
            \STATE Receive global FL parameters $\left\{\boldsymbol{w}_k^G\left(t\right)| k\in\mathcal{K}\right\}$;
                % \FOR{$k = 1, 2, \hdots, K$}
                %     \STATE Calculate loss $\mathcal{L}_{i, k}^{t}$ by (\ref{eq:training_loss}).
                %     \STATE Obtain gradient $\nabla\mathcal{L}_{i, k}^{t}$.
                %     \STATE Calculate gradient similarity $S_{i, k}\left(t\right)$ by (\ref{eq:similarity}).
                % \ENDFOR
                \STATE Calculate loss $\mathcal{L}_{i, 1}^{t}, \ldots, \mathcal{L}_{i, K}^{t}$ by (\ref{eq:training_loss});
                \STATE Obtain gradient $\nabla\mathcal{L}_{i, 1}^{t}, \dots, \nabla\mathcal{L}_{i, K}^{t}$;
                \STATE Calculate gradient similarity $S_{i, 1}\!\!\left(t\right), \ldots, S_{i, K}\!\!\left(t\right)$ by (\ref{eq:similarity});
                \STATE Estimate task identity $\epsilon_i\left(t\right)$ by (\ref{eq:identity});
                \STATE Update the local model $\boldsymbol{w}^i_{\epsilon_i\left(t-1\right)}\left(t\right)$ by (\ref{eq:update});
                \STATE Broadcast $\epsilon_i\left(t\right)$ to all BSs;
            \ENDFOR
            \STATE \underline{BS:} determine its associated users based on user task identities;
            \STATE \underline{Each user $i$:} upload its local models;
            \STATE \underline{BS:} aggregate local models from its associated users by (\ref{eq:updated_global});
            \STATE \underline{BS:} share its models with other BSs to generate common FL global models by (\ref{eq:get_global});

        \ENDFOR

        \STATE \textbf{Output:} global task models: $\left\{\boldsymbol{w}_{k}^G\left(T\right)|\forall~b~\in~\mathcal{B},~\forall k~\in~\mathcal{K}_b\right\}$ and the task identities of users: $\left\{\epsilon_1\left(T\right), \epsilon_2\left(T\right), \hdots, \epsilon_U\left(T\right)\right\}$;
    \end{algorithmic}
    \vspace{-0.1cm}
\end{algorithm}
\vspace{-0.2cm}

\subsection{Data Transmission Model}

\emph{1) Local FL parameter transmission:}
The orthogonal frequency division multiple access (OFDMA) protocol is used for local FL model parameter transmission from users to BSs \cite{tan2021energy}. 
We assume that each user can only occupy one RB, and the BSs can allocate the same set $\mathcal{R}$ of $R$ RBs to the users. 
The uplink rate of user $i$ transmitting its FL model parameters to BS~$b$ is
\begin{equation} \label{eq:uplink}
    c_{i,b,t}^\textrm{U}\left(\boldsymbol{r}_{i,b}\left(t\right)\right) =
    \sum_{n=1}^R r_{i,b}^n\left(t\right) B^\textrm{U} \log_2\left(1 + \frac{P_i h_{i,b}}{I^n_{i,b}\left(t\right) + B^\textrm{U} W_0}\right),
\end{equation}
where $\boldsymbol{r}_{i,b}\left(t\right) = \left[r_{i,b}^1\left(t\right), \ldots, r_{i,b}^R\left(t\right)\right]$ is an RB allocation vector with $r_{i,b}^n\left(t\right) \in \{0,1\}$: $r_{i,b}^n\left(t\right) = 1$ implying that RB $n$ is allocated to user $i$ at time slot $t$; otherwise, we have $r_{i,b}^n\left(t\right) = 0$. 
% So we have
% \begin{equation}\label{eq:r_a}
%     \sum_{n=1}^Rr_{i,b}^n\left(t\right) = a_{i,b}\left(t\right),
% \end{equation}
% We define $\sum_{n=1}^Rr_{i,n} = a_i$, where $a_i = 1$ means that the user $i$ is allocated RB and FL algorithm is performed successfully, while $a_i = 0 $ is otherwise.
$B^\textrm{U}$ is the bandwidth of each RB and $h_{i,b}$ is the channel gain from user $i$ to BS $b$; $W_0$ is the noise power spectral density; $I^n_{i,b}\left(t\right) = \sum_{p \in \mathcal{U} ,p \neq i} \limits \sum_{q \in \mathcal{B} , q \neq b} \limits r_{p,q}^n\left(t\right)P_p h_{p,q}$ is the inter-user interference caused by the users that use the same RB $n$ but connect to other BSs. 

\textcolor{black}{\emph{2) Global FL model parameter transmission:}
\textcolor{black}{For global FL model parameter transmission from BSs to users, the time division multiple access (TDMA) is applied \cite{deng2020implementing}.}
Let $W_k^G\left(t\right)$ be the data size of the global FL model $\boldsymbol{w}_k^G\left(t\right)$, and $c^{\textrm{D}}_t$ be the downlink data rate of each BS $b$.
The delay of common FL parameter transmission from BS $b$ to all users is
\begin{equation}
    D_b\left(t\right) = \frac{\sum\limits_{k \in \mathcal{K}_b} \mathds{1}_{\big\{\sum\limits_{i\in \mathcal{U}_{k,t}} a_{i,b}\left(t\right) > 0\big\}}W_k^G\left(t\right)}{c^{\textrm{D}}_t},
\end{equation}
where $\mathds{1}_{\big\{\sum\limits_{i\in \mathcal{U}_{k,t}} a_{i,b}\left(t\right) > 0\big\}}$ is a function that indicates whether BS $b$ aggregates FL models for task $k$. 
Specifically, $\mathds{1}_{\big\{\sum\limits_{i\in \mathcal{U}_{k,t}} a_{i,b}\left(t\right) > 0\big\}} = 1$  indicates that the users with task identity $k$ is connected to BS $b$, BS $b$ must aggregate FL model parameters for task $k$ and broadcast the global model to all users.
Otherwise, we have $\mathds{1}_{\big\{\sum\limits_{i\in \mathcal{U}_{k,t}} a_{i,b}\left(t\right) > 0\big\}} = 0$.}

\subsection{Problem Formulation}
Given the defined models, our goal is to minimize the training loss of $K$ FL models while considering data privacy and FL convergence time. 
This minimization problem involves optimizing the RB allocation strategy for each user as well as minimizing the privacy leakage during FL training.  
The minimization problem is formulated as
\begin{equation} \label{eq:problem_formulation}
    \begin{aligned}
        \min_{\substack{\boldsymbol{r}_{i,b}\left(t\right),\\ a_{i,b}\left(t\right), \\ \sigma_i}} \quad \frac{1}{\sum_{i=1}^{U}X_i} & \sum_{i=1}^{U} \left( \sum_{k=1}^{X_i} l\left( \boldsymbol{w}^i_{\epsilon_i\left(T\right)}\left(T\right), \boldsymbol{x}_{i,k}, y_{i,k} \right) \right. \\
        & \left. + \gamma \left( \frac{1}{X_i\sigma_i} \right)^2 \sum_{b=1}^{B} a_{i,b}\left(T\right) \right),
    \end{aligned}
\end{equation}

% manually tag equations
\begin{align} 
    \text{s.t.} \quad & a_{i,b}\left(t\right), r_{i,b}^n\left(t\right) \!\in\! \{0,1\}, \forall i \!\in\! \mathcal{U}, \forall n \in \mathcal{R}, \forall b \in \mathcal{B}, \tag{11a} \label{eq:constraint_a} \\
    & a_{i,b}\left(t\right) = \sum_{n=1}^Rr_{i,b}^n\left(t\right), \sum_{i=1}^Ua_{i,b}\left(t\right) \leq R, \forall b \in \mathcal{B}, \tag{11b} \label{eq:constraint_b} \\
    & \sum_{b=1}^BD_b\left(t\right) \leq \gamma_D, \tag{11c} \label{eq:constraint_c} \\
    &c_{i,b,t}^\textrm{U}(\boldsymbol{r}_{i,b}\left(t\right)) \geq \gamma_T, \forall i \in \mathcal{U}, \forall b \in \mathcal{B}, \tag{11d} \label{eq:constraint_d} \\
    & \begin{aligned}
    &\sum_{b\in\mathcal{B}} \sum_{i\in\mathcal{U}_{k,T}} X_i \sigma_{i}^2 \sum_{n=1}^R r_{i,b}^n\left(t\right) \leq \\ 
    &V_{\textrm{max}}\sum_{b\in\mathcal{B}} \sum_{i\in\mathcal{U}_{k,T}} X_i \sum_{n=1}^R r_{i,b}^n\left(t\right), \quad \forall k \in \mathcal{K}, 
    \end{aligned} \tag{11e} \label{eq:constraint_e} \\
    & X_i\sigma_{i} \geq N_{\textrm{min}}\sum_{n=1}^Rr_{i,b}^n\left(t\right), \forall b\in \mathcal{B}, \forall i\in \mathcal{U}, \tag{11f} \label{eq:constraint_f}
\end{align}
where $\gamma > 0$ is a constant that is used to balance the FL training loss and the total privacy leakage; $\gamma_T$ is the data rate requirement of FL model parameter transmission; $\gamma_D$ is the delay requirement of FL model parameter transmission; $V_{\textrm{max}}$ and $N_{\textrm{min}}$ are two constants that are used to control the DP noise \cite{tavangaran2024differential}. Constraints (\ref{eq:constraint_a}) and (\ref{eq:constraint_b}) indicate that each user can occupy only one RB for FL model parameter transmission. Constraint (\ref{eq:constraint_c}) is the data rate requirement of local FL model transmission from each user to its associated BS.
Constraint (\ref{eq:constraint_d}) is the delay requirement of global FL model transmission from each BS to the users.
Constraint (\ref{eq:constraint_e}) guarantees that the DP noise error is less than a given constant $V_{\textrm{max}}$.
% We can control the amount of DP noise variance and its error by adjusting the constant $V_{\textrm{max}}$.
Constraint (\ref{eq:constraint_f}) indicates the minimum DP noise that must be added to each user. 
% Both (\ref{eq:constraint_e}) and (\ref{eq:constraint_f}) set up the requirements to guarantee the quality of DP noise error.
% Since problem (\ref{eq:problem_formulation}) is a non-convex mixed-integer optimization problem, which is generally challenging to solve, we propose employing a genetic algorithm to tackle this problem.

\textcolor{black}{The problem in (\ref{eq:problem_formulation}) is challenging to solve due to the following reasons.
First, the BSs cannot verify the user task identity estimated by themselves.
The task identification estimation result will affect the BSs' decisions of user association and RB allocation. Meanwhile, there is no explicit equations to capture the relationship between the total loss function and user task identity, user association, and RB allocation.
Second, the DP parameter $\boldsymbol{n}_i\left(t\right)$, user association variable $\boldsymbol{a}_{i,b}\left(t\right)$, and RB allocation vector $\boldsymbol{r}_{i,b}\left(t\right)$ affect not only the local FL model transmission but also the global FL transmission delay $D_b\left(t\right)$ and data privacy leakage. This is different from the prior works \cite{tran2019federated, 9210812} that only consider the optimization of RB allocation for local FL model transmission. 
Third, in problem (\ref{eq:problem_formulation}), we consider a scenario where several BSs jointly execute FL algorithms and they must determine user association and RB allocation simultaneously. 
Hence, each BS cannot know the strategies of user association and RB allocation of other BSs, which increases the complexity of solving the problem. }
% To tackle these challenges, we propose employing a value decomposed multi-agent reinforcement learning (DPVD-MARL) algorithm to solve this problem. 
% Compared with traditional centralized learning solutions, DPVD-MARL makes each BS determine its RB allocation strategies and users individually manage DP noise parameters that will be trained and transmitted.
% Hence, with DPVD-MARL, the BSs and users can jointly minimize the final loss and the privacy leakage while meeting the transmission and noise energy constraints.

\section{Analysis of the CFL Convergence Rate}

To solve problem (\ref{eq:problem_formulation}), we first analyze the convergence of our proposed CFL algorithm for both convex and non-convex loss functions. The convergence analytical result will be instrumental for the design of the solution algorithm.
Let $\mathcal{T}_k$ be the set of $T_k$ users that should perform task $k$.
However, these users may be incorrectly clustered into other task groups due to distributed clustering methods in (\ref{eq:identity}).
Let \( p_{k, i}^t \) be the probability that user $i$ from task \( k \) is incorrectly clustered into any other groups at iteration $t$. 
We assume that the probability that user $i$ is identified into each incorrect task is equal.  
For example, $\frac{1}{K-1}p_{k, i}^t$ is the probability that user $i$ should participate in the FL training of task $k$ but it identifies itself as a user of task $k'$.

\subsection{Convergence Analysis of CFL with Convex Loss Functions}

Let $F\left(\boldsymbol{w}_{k}^G\left(t\right)\right) = \frac{1}{\sum\limits_{\substack{i \in \mathcal{T}_k}} X_i} \sum\limits_{\substack{i \in \mathcal{T}_k}} \sum\limits_{\substack{m=1}}^{X_i} l\left(\boldsymbol{w}_k^G\left(t\right), \boldsymbol{x}_{i,m}, y_{i,m}\right)$ be the loss of all users of task $k$. Let $a_i\left(t\right) = \mathds{1}_{\left\{\sum\limits_{b=1}^B a_{i,b}\left(t\right) > 0\right\}}$ be the indicator weather user $i$ is associated with a BS. Based on (\ref{eq:updated_global}) and (\ref{eq:get_global}), we can rewrite the update of the global FL model of task $k$ at step $t$ as
\begin{equation} \label{eq:global_upgrade}
    \boldsymbol{w}_k^G\left(t+1\right) = \boldsymbol{w}_k^G\left(t\right) - \alpha\left(\nabla F\left(\boldsymbol{w}_{k}^G\left(t\right)\right) - \boldsymbol{o}_k + \boldsymbol{e}_k + \boldsymbol{n}_k\right),
\end{equation}
where
\begin{equation}
\boldsymbol{o}_k = \nabla F\left(\boldsymbol{w}_{k}^G\left(t\right)\right) - \frac{\sum\limits_{\substack{i \in \mathcal{T}_k}}a_i\left(t\right)\sum\limits_{\substack{m=1}}^{X_i}\nabla l_{i,m}\left(\boldsymbol{w}_k^G\left(t\right)\right)}{\sum\limits_{\substack{i \in \mathcal{T}_k}}X_ia_i\left(t\right)},
\end{equation}
\begin{equation}
\boldsymbol{e}_k \!= \!\frac{\sum\limits_{\substack{i \in \mathcal{U}_{k,t}}}\!\!\!\!a_i\left(t\right)\!\!\!\sum\limits_{{m=1}}^{X_i}\!\!\nabla l_{i,m}\!\!\left(\boldsymbol{w}_k^G\left(t\right)\right)}{\sum\limits_{\substack{i \in \mathcal{U}_{k,t}}}X_ia_i\left(t\right)}\!
- \!\frac{\sum\limits_{{i \in \mathcal{T}_k}}\!\!a_i\left(t\right)\!\!\!\sum\limits_{\substack{m=1}}^{X_i}\!\!\nabla l_{i,m}\!\!\left(\boldsymbol{w}_k^G\left(t\right)\right)}{\sum\limits_{{i \in \mathcal{T}_k}}X_ia_i\left(t\right)},
\end{equation}
and
\begin{equation}
\boldsymbol{n}_k = \frac{1}{\sum\limits_{i \in \mathcal{U}_{k,t}}X_ia_i\left(t\right)}\sum\limits_{i \in \mathcal{U}_{k,t}}X_ia_i\left(t\right)\boldsymbol{n}_i\left(t\right),
\end{equation}
with $\nabla l_{i,m}\left(\boldsymbol{w}_k^G\left(t\right)\right)$ being short for $\nabla l\left(\boldsymbol{w}_k^G\left(t\right), \boldsymbol{x}_{i,m}, y_{i,m}\right)$.

\noindent From (\ref{eq:global_upgrade}), we see that $\boldsymbol{o}_k$ is the error introduced by users that belong to task $k$ but do not join the training process of task $k$. $\boldsymbol{e}_k$ is the error caused by the users that are incorrectly identified as users in cluster $k$ and join the training process of task $k$.
$\boldsymbol{n}_k$ is the DP noise error of the users that perform task $k$.
We also assume that the CFL algorithm converges to an optimal global FL model $\boldsymbol{w}^*_k$ after a certain number of learning steps. To derive the expected convergence rate of CFL, we first make the following assumptions.
\begin{itemize}
    \item We assume that the gradient $\nabla F\left(\boldsymbol{w}_{k}^G\left(t\right)\right)$ of $F\left(\boldsymbol{w}_{k}^G\left(t\right)\right)$ is uniformly Lipschitz continuous with respect to $\boldsymbol{w}_{k}^G\left(t\right)$. Hence, we have
    \begin{equation} \label{eq:lips}
        \left\|\!\nabla\! F\!\!\left(\!\boldsymbol{w}_{k}^G\!\!\left(t\!\!+\!\!1\right)\!\right) \!\!-\!\! \nabla\! F\!\!\left(\boldsymbol{w}_{k}^G\left(t\right)\right)\!\right\|\!\! \leq \!\!L\!\left\|\!\boldsymbol{w}_{k}^G\!\!\left(\!t\!\!+\!\!1\!\right)\!\! - \!\!\boldsymbol{w}_{k}^G\!\!\left(t\right)\!\right\|\!,
    \end{equation}
    where $L$ is a positive constant and $\left\|\boldsymbol{w}_{k}^G(t+1) - \boldsymbol{w}_{k}^G\left(t\right)\right\|$ is the norm of $\boldsymbol{w}_{k}^G(t+1) - \boldsymbol{w}_{k}^G\left(t\right)$.
    \item We assume that $F\left(\boldsymbol{w}_{k}^G\left(t\right)\right)$ is strong convex with positive parameter $\mu$, such that
    \begin{equation} \label{eq:convex}
    \begin{aligned}
        F\left(\boldsymbol{w}_{k}^G\left(t+1\right)\right) &\geq F\left(\boldsymbol{w}_{k}^G\left(t\right)\right)\\
        &+ \left(\boldsymbol{w}_{k}^G(t+1) \!\!-\!\! \boldsymbol{w}_{k}^G\left(t\right)\right)^{T}\!\!\nabla\! F\!\left(\!\boldsymbol{w}_{k}^G\left(t\right)\!\right)\\
        &+ \frac{\mu}{2}\left\|\boldsymbol{w}_{k}^G\left(t+1\right) - \boldsymbol{w}_{k}^G\left(t\right)\right\|^2.
    \end{aligned}
    \end{equation}
    \item We assume that $F\left(\boldsymbol{w}_{k}^G(t+1)\right)$ is twice-continuously differentiable. Based on (\ref{eq:lips}) and (\ref{eq:convex}), we have
    \begin{equation} \label{eq:twice}
        \mu \boldsymbol{I} \preceq \nabla^2 F\left(\boldsymbol{w}_{k}^G\left(t\right)\right) \preceq L \boldsymbol{I}.
    \end{equation}
    \item We assume that, with $\zeta_1, \zeta_2 \geq 0$, we have
    \begin{equation} \label{eq:zeta}
        \left\|\nabla l\left(\boldsymbol{w}_k^G\left(t\right), \boldsymbol{x}_{i,m}, y_{i,m}\right)\right\|^2\!\! \leq \zeta_1 \!+\! \zeta_2 \left\|\nabla F\left(\boldsymbol{w}_{k}^G\left(t\right)\right)\right\|^2\!\!.
    \end{equation}
    \textcolor{blue}{In \eqref{eq:zeta}, $\zeta_2$ captures the variation of the local gradients with respect to the global gradient. When $\zeta_2$ is small, the local gradients are close to the global gradient direction.}
\end{itemize}
These assumptions can be satisfied by several widely used loss functions such as the mean squared error, logistic regression, and cross entropy. These popular loss functions can be used to capture the performance of implementing practical CFL algorithms for identification, classification, and prediction. 
% Let \( A_k \) be the lower bound on the total number of data samples from users that the BSs connect with in task \( k \), and $A_k$ follows:
% \begin{equation} \label{eq:ak}
%     A_k \leq \sum\limits_{i \in \mathcal{U}_k}X_ia_i\left(t\right), \forall k\in \mathcal{K}.
% \end{equation}
% The expected convergence rate of the CFL algorithms can now be derived by the following theorem.
\begin{theorem}\label{the:1}
Given the user selection vector $\boldsymbol{a}$, optimal global FL model $\boldsymbol{w}^*_k$ of task $k$, and the learning rate $\alpha = \frac{1}{L}$, $\mathbb{E}\left(F\left(\boldsymbol{w}_{k}^G\left(t+1\right)\right) - F\left(\boldsymbol{w}_{k}^*\right)\right)$ is upper bounded as follows:
\begin{equation} \label{eq:theorem1}
\begin{aligned}
    &\mathbb{E} \left( F\left(\boldsymbol{w}_k^G(t+1)\right) - F\left(\boldsymbol{w}_k^*\right) \right) \\
    &\leq \frac{2\zeta_1Q_{1,k,t}}{L- \mu + 4 \mu \zeta_2}\!+\! \frac{d^2\sigma_{max}^2}{2L} \!+ \!Q_{1,k,t}\left(F\left(\boldsymbol{w}_k^G\left(t\right)\right)\! -\! F\left(\boldsymbol{w}_k^*\right)\right)\!,
\end{aligned}
\end{equation}
where $\mathbb{E}(\cdot)$ is the expectation with respect to the incorrect task clustering and DP noise of the users, and
\begin{equation} \label{eq:q1kt}
\begin{aligned}
    Q_{1,k,t} &= \frac{L- \mu + 4 \mu \zeta_2}{L}\\
    &\quad\times\left(1 - \frac{1}{\sum\limits_{i \in \mathcal{T}_k}X_i}\sum\limits_{i \in \mathcal{T}_k}X_i\left(1-p_{k,i}^t\right)a_i\left(t\right)\right).
\end{aligned}
\end{equation}
\end{theorem}
\noindent\textbf{Proof.}
See Appendix 1.

In Theorem \ref{the:1}, $\boldsymbol{w}^*_k$ is the optimal FL model that is generated based on local FL models of users of $\mathcal{T}_k$ under an ideal setting where all users are classified correctly (i.e., $\sum\limits_{k' \neq k}p_{k',i}^{t} = 0$) and all users are connected to the BSs (i.e., $\prod\limits_{i \in \mathcal{T}_k}a_i\left(t\right) = 1$).
From Theorem \ref{the:1}, we see that a gap, $\frac{2\zeta_1Q_{1,k,t}}{L- \mu + 4 \mu \zeta_2}+ \frac{d^2\sigma_{max}^2}{2L}$, exists between $\mathbb{E}\left(F\left(\boldsymbol{w}_k^G\left(t+1\right)\right)\right)$ and $\mathbb{E}\left(F\left(\boldsymbol{w}_k^G\left(t\right)\right)\right)$. This gap is caused by the incorrect clustering probabilities $p_{k,i}^t$ and the DP noise $\sigma_i$ from selected users. As the incorrect classification probabilities of users $p_{k,t}^t$ decrease, $Q_{1,k,t}$ also decreases. Meanwhile, if BSs select more users to connect to (i.e., $\sum\limits_{i \in \mathcal{T}_k}a_i$ increases), $Q_{1,k,t}$ will also decrease. Moreover, selecting users with more data samples also reduces $Q_{1,k,t}$. As the gap decreases, the convergence speed of the CFL algorithm improves. Hence, it is necessary to optimize resource allocation, user selection, and DP noise for the implementation of CFL algorithms.

Based on Theorem \ref{the:1}, we also analyze the value of $\zeta_2$ to guarantee the convergence of CFL algorithms, as shown in the following proposition.

\begin{proposition} \label{pro:1}
    Given the learning rate $\alpha = \frac{1}{L}$, to guarantee the CFL convergence in (\ref{eq:theorem1}), $\zeta_2$ must satisfy
    \begin{equation} \label{eq:zeta_2}
        0 < \zeta_2 < \frac{1}{4}.
    \end{equation}
\end{proposition}
\noindent\textbf{Proof.}
Let $Q_{1,k,max} = \max\left(Q_{1,k,0}, Q_{1,k,1}, \ldots, Q_{1,k,t}\right)$. From Theorem \ref{the:1}, we see that when $Q_{1,k,max} < 1$, $Q_{1,k,max}^t$, the $t$-th power of $Q_{1,k,max}$, is equal to $0$. Hence, 
we have
\begin{equation} \label{eq:multiple}
\begin{aligned}
    &\mathbb{E} \left( F\left(\boldsymbol{w}_k^G\left(t+1\right)\right) - F\left(\boldsymbol{w}_k^*\right) \right) \\
    &\leq \frac{2\zeta_1}{L- \mu + 4 \mu \zeta_2}\sum\limits_{k=0}^{t-1}Q_{1,k,max}^{k+1}+ \frac{d^2\sigma_{max}^2}{2L}\sum\limits_{k=0}^{t-1}Q_{1,k,max}^{k+1} \\
    & \quad + Q_{1,k,max}^t \mathbb{E} \left( F\left(\boldsymbol{w}_k^G\left(0\right)\right) - F\left(\boldsymbol{w}_k^*\right) \right) \\
    % &\quad + \frac{d^2\sigma_{max}^2}{2L}\sum\limits_{k=0}^{t-1}Q_{1,k,max}^{k+1}\\
    & =\frac{2\zeta_1}{L- \mu + 4 \mu \zeta_2}\frac{Q_{1,k,max}-Q_{1,k,max}^{t+1}}{1-Q_{1,k,max}} \\
    & \quad + \frac{d^2\sigma_{max}^2(1-Q_{1,k,max}^t)}{2L\left(1-Q_{1,k,max}\right)}\\
    & \quad + Q_{1,k,max}^t \mathbb{E} \left( F\left(\boldsymbol{w}_k^G\left(0\right)\right) - F\left(\boldsymbol{w}_k^*\right) \right) \\
    & = \frac{2\zeta_1}{L- \mu + 4 \mu \zeta_2}\frac{Q_{1,k,max}}{1-Q_{1,k,max}} + \frac{d^2\sigma_{max}^2}{2L\left(1-Q_{1,k,max}\right)}.
\end{aligned}
\end{equation}
% $\mathbb{E} \left( F(\boldsymbol{w}_k^G(t+1)) - F(\boldsymbol{w}_k^*) \right) = \frac{2\zeta_1}{L- \mu + 4 \mu \zeta_2}\frac{Q_{2,t}}{1-Q_{2,t}} + \frac{Q_1}{2L(1-Q_{2,t})}$ 
From (\ref{eq:multiple}), we can see that, to guarantee the CFL convergence, we only need to make $Q_{1,k,t} = \frac{L- \mu + 4 \mu \zeta_2}{L}Q_{0,k,t} < 1, \forall i \in \left[0,t\right]$, where $Q_{0,k,t}$ is generated in (49).
Since $0\leq Q_{0,k,t} \leq 1$, we only need to ensure that $4\mu \zeta_2 - \mu < 0$.
Therefore, we have $\zeta_2 < \frac{1}{4}$.
Since $\zeta_2$ must satisfy $\left\|\nabla l(\boldsymbol{w}_k^G\left(t\right), \boldsymbol{x}_{i,m}, y_{i,m})\right\|^2 \leq \zeta_1 + \zeta_2 \left\|\nabla F\left(\boldsymbol{w}_{k}^G\left(t\right)\right)\right\|^2$, we have $\zeta_2 > 0$. This completes the proof.

\textcolor{blue}{Proposition~\ref{pro:1} provides a sufficient condition to ensure CFL convergence. It provides a theoretical guideline for selecting loss functions and understanding how gradient variance affects training behavior. Here, if a loss function cannot meet the condition in Proposition 1, the designed CFL may not converge, which must be further evaluated by simulations. In contrast, if a loss function satisfies the condition, the designed CFL will converge. }

\subsection{Convergence Analysis of CFL with Non-convex Loss Functions}

Next, we extend the convergence analysis of CFL in Theorem \ref{the:1} to that of  the proposed CFL with a non-convex loss function. In particular, we first replace convex assumptions (\ref{eq:convex}) and (\ref{eq:twice}) with the following assumptions:

\begin{itemize}
    \item We assume that the gradient of the non-convex loss function $F\left(\boldsymbol{w}_{k}^G\!\left(t\right)\right)$ has an upper bound, i.e., $\left\|\nabla F\left(\boldsymbol{w}_{k}^G\!\left(t\right)\right)\right\| \leq C$ \cite{fallah2020personalized}.
    \item We assume that the function $F\left(\boldsymbol{w}_{k}^G\!\left(t\right)\right)$ is $\mu$-non-convex such that all eigenvalues of $\nabla^2 F\left(\boldsymbol{w}_{k}^G\!\left(t\right)\right)$ lie in $\left[-\mu, L\right]$, for some $\mu \in \left(0,L\right]$ \cite{wang2022uav}.
\end{itemize}
% \textbf{Condition 1}\cite{fallah2020personalized}: The gradient of the non-convex loss function $F\left(\boldsymbol{w}_{k}^G\!\left(t\right)\right)$ has an upper bound, i.e., $\left\|\nabla F\left(\boldsymbol{w}_{k}^G\!\left(t\right)\right)\right\| \leq C$.
% \textbf{Condition 2}\cite{wang2022uav}: Function $F\left(\boldsymbol{w}_{k}^G\!\left(t\right)\right)$ is $\mu$-non-convex such that all eigenvalues of $\nabla^2 F\left(\boldsymbol{w}_{k}^G\!\left(t\right)\right)$ lie in $\left[-\mu, L\right]$, for some $\mu \in \left(0,L\right]$ .
% \textbf{Condition 3}\cite{wang2022uav}: The Hessian of the loss function $F\left(\boldsymbol{w}_{k}^G\!\left(t\right)\right)$ is $\gamma$-Lipschitz continuous, such that 
% \begin{equation} \label{eq:mu_lip}
%     \left\|\nabla^2 F\left(\boldsymbol{w}_{k}^G\left(t\right)\right)-\nabla^2 F\left(\boldsymbol{w}_{k}^G\left(t'\right)\right)\right\| \leq \gamma \left\|\boldsymbol{w}_{k}^G\left(t\right)-\boldsymbol{w}_{k}^G\left(t'\right)\right\|.
% \end{equation}
Based on the above assumptions, the convergence of our proposed CFL method with non-convex loss functions is presented in the following theorem.

\begin{theorem}\label{the:2}
Given the user selection vector $\boldsymbol{a}$, optimal global FL model $\boldsymbol{w}^*_k$ of task $k$, and the learning rate $\alpha = \frac{1}{L}$, $\mathbb{E}\left(F\left(\boldsymbol{w}_{k}^G\left(t+1\right)\right) - F\left(\boldsymbol{w}_{k}^*\right)\right)$ is upper bounded as follows:
\begin{equation} \label{eq:theorem2}
\begin{aligned}
    &\mathbb{E} \left( F\left(\boldsymbol{w}_k^G\left(t+1\right)\right) - F\left(\boldsymbol{w}_k^*\right) \right) \\
    &\leq Q_{1,k,t}\left(F\left(\boldsymbol{w}_k^G\left(t\right)\right) \!-\! F\left(\!\boldsymbol{w}_k^*\!\right)\right) \\
    & \quad +\frac{2\zeta_1Q_{1,k,t}}{L- \mu + 4 \mu \zeta_2}+\frac{d^2\sigma_{max}^2+\eta \left(1-4 \zeta_2\right)Q_{0,k,t}}{2L},
\end{aligned}
\end{equation}
\end{theorem}
\noindent\textbf{Proof.}
With the assumption of the values of eigenvalues of $\nabla^2 F\left(\boldsymbol{w}_{k}^G\!\left(t\right)\right)$, (42) in Theorem \ref{the:1} still holds. According to \cite{vlaski2021distributed} and \cite{mulvaney2022dynamic}, as well as the above assumptions, (54) in Theorem \ref{the:1} can be rewritten as
\begin{equation} \label{eq:nonononono}
    \left\|\nabla F\left(\boldsymbol{w}_k^G\left(t\right)\right)\right\|^2 \geq 2\mu\left(F\left(\boldsymbol{w}_k^G\left(t\right)\right) - F\left(\boldsymbol{w}_k^*\right)\right) - \eta,
\end{equation}
where $\eta$ depends on the magnitude of the negative eigenvalues of $\nabla^2 F\left(\boldsymbol{w}_{k}^G\!\left(t\right)\right)$.
Then, we use the method used to obtain (55) to find the upper bound of $\mathbb{E} \left( F\left(\boldsymbol{w}_k^G\left(t+1\right)\right) - F\left(\boldsymbol{w}_k^*\right) \right)$, which is given by
\begin{equation} \label{eq:one_time_2}
\begin{aligned}
    &\mathbb{E} \left( F\left(\boldsymbol{w}_k^G\left(t+1\right)\right) - F\left(\boldsymbol{w}_k^*\right) \right) \\
    &\leq Q_{1,k,t}\left(F\left(\boldsymbol{w}_k^G\left(t\right)\right) \!-\! F\left(\!\boldsymbol{w}_k^*\!\right)\right) \\
    & \quad +\frac{2\zeta_1Q_{1,k,t}}{L- \mu + 4 \mu \zeta_2}+\frac{d^2\sigma_{max}^2+\eta \left(1-4 \zeta_2\right)Q_{0,k,t}}{2L},
\end{aligned}
\end{equation}
This completes the proof.

%From Theorem \ref{the:2}, we see that the CFL convergence rate depends on the user participation strategy. Given the above assumptions, we are able to limit the impact of $\eta$ and still find the optimal scheme that ensures the convergence of the global FL model.

\section{Dynamic Penalty Function Assisted Value Decomposition Based Multi-agent RL}
\label{Value Decomposition Based Multi-agent RL}

To solve the problem in (\ref{eq:problem_formulation}), we cannot directly use the convergence analytical result in Theorem \ref{the:1}, as done in several existing works \cite{tavangaran2024differential}, where (\ref{eq:theorem1}) includes known parameters $p_{k,i}^t$. To solve problem (\ref{eq:problem_formulation}), %and address the aforementioned challenges, 
we propose a novel dynamic penalty function assisted value decomposed multi-agent reinforcement learning (DPVD-MARL) algorithm with joint optimization method. 
DPVD-MARL enables each BS to determine its user association, RB allocation strategies, and DP noise parameters, such that they can jointly optimize the total loss and privacy leakage in (\ref{eq:problem_formulation}).
In the proposed method, we use a novel penalty assignment scheme, which assigns penalty depending on the number of devices that cannot meet communication constraints (e.g., delay), to guide the DPVD-MARL to quickly find valid actions, thus improving the convergence speed.
We first introduce the components of the proposed method and then elaborate on its training process.

\subsection{Components of DPVD-MARL}
In this section, we introduce the fundamental components of DPVD-MARL as follows:

    \textcolor{black}{\emph{1) Agents:} The agents performing DPVD-MARL are the BSs. Each BS determines its RB allocation, user association, and DP noise parameters.}
    
    \emph{2) States:} The state of each BS $b$ is used to describe the current wireless environment and the FL training status, represented by $\boldsymbol{s}_b\left(t\right) = \left[\boldsymbol{\epsilon}_b\left(t\right), \boldsymbol{d}_b\left(t\right)\right]$, with $\boldsymbol{\epsilon}\left(t\right) = \left[\epsilon_1\left(t\right), \ldots, \epsilon_U\left(t\right)\right]$ being the task identity vector of all users, and
    $\boldsymbol{d}_b\left(t\right) = \left[d_{b,1}\left(t\right), \ldots, d_{b,U}\left(t\right)\right]$ being the distance vector between BS $b$ and each user.
    The global state at iteration $t$ is $\boldsymbol{s}\left(t\right) = \left[\boldsymbol{s}_{1}\left(t\right), \ldots, \boldsymbol{s}_{B}\left(t\right)\right]$.
    
    \emph{3) Actions:} The action of each BS $b$ is to determine if it connects to each user. Let $\boldsymbol{a}_b\left(t\right) = \left[a_{1, b}\left(t\right), \ldots, a_{U, b}\left(t\right)\right]$ be an action of BS $b$  at iteration $t$. 
    Here, each action does not include RB allocation strategy $r_{i,b}^n\left(t\right)$ and DP noise parameter $\sigma_i$, since both of these parameters can be determined via a joint optimization method when the action vector $\boldsymbol{a}_b\left(t\right)$ is determined.  
    Then, $\boldsymbol{a}\left(t\right) = \left[\boldsymbol{a}_1\left(t\right), \ldots,  \boldsymbol{a}_B\left(t\right)\right]$ is the action vector of all the BSs. 
    
    \emph{4) Shared reward:} The shared reward of all BSs captures the training loss of the learning tasks implemented by all BSs and the data privacy leakage of all users. Given the global state $\boldsymbol{s}\left(t\right)$ and the selected action $\boldsymbol{a}\left(t\right)$, the shared reward for all the BSs at iteration $t$ is
    \begin{equation} \label{eq:shared_reward}
    \begin{aligned}
        & R\left(\boldsymbol{a}\left(t\right)|\boldsymbol{s}\left(t\right)\right) = -\frac{1}{U}\sum_{i=1}^U \Bigg( \sum_{k=1}^{X_i} l\left( \boldsymbol{w}^{i}_{\epsilon_i\left(t\right)}\left(t\right), \boldsymbol{x}_{i,k}, y_{i,k} \right) \\
        & \quad + \mathds{1}_{\left\{\sum_{b=1}^B a_{i,b}\left(t\right) > 0\right\}} \gamma \left( \frac{1}{X_{i}\sigma_{i}} \right)^2 \Bigg)
        - \left(\sum_{i=1}^U \Phi_i \right)^2 \times 10^3,
    \end{aligned}
    \end{equation}
    where $\Phi_i$ is the conditional function for each user $i$, which is %given by
    \begin{equation} \label{eq:phi_i}
    \begin{aligned}
    &\Phi_i = 
    \begin{cases} 
        0, \qquad \textrm{if condition (\ref{eq:constraint_a}) to (\ref{eq:constraint_c}) are satisfied,} &  \\
        1, \qquad  \textrm{otherwise.} & 
    \end{cases}
    \end{aligned}
    \end{equation}
    % \begin{equation} \label{eq:reward}
    % \begin{aligned}
    % &R(\boldsymbol{a}\left(t\right)|\boldsymbol{s}\left(t\right)) = \\
    % &\begin{cases} 
    %     -\sum_{i=1}^U \left( \sum_{k=1}^{X_i} l\left( \boldsymbol{w}^{i}_{\epsilon_i\left(t\right)}\left(t\right), \boldsymbol{x}_{i,k}, y_{i,k} \right) + \mathds{1}_{\{\sum_{b=1}^Ba_{i,b}\left(t\right) > 0\}}\right. \\
    %     \left. \gamma \left( \frac{1}{X_{i}\sigma_{i}} \right)^2 \right) \qquad \textrm{if condition (\ref{eq:constraint_a}) to (\ref{eq:constraint_d}) are satisfied,} &  \\
    %     -10^6 \qquad \qquad \qquad \textrm{otherwise.} & 
    % \end{cases}
    % \end{aligned}
    % \end{equation}
    From (\ref{eq:shared_reward}) and (\ref{eq:phi_i}), we can see that only when BSs connects to its associated users (i.e., $a_{i,b}\left(t\right) = 1$) and all constraints (\ref{eq:constraint_a}) to (\ref{eq:constraint_c}) are satisfied, the reward will be \begin{equation} \label{eq:reward_satisfy}
    \begin{aligned}
        R\left(\boldsymbol{a}\left(t\right)|\boldsymbol{s}\left(t\right)\right) &= 
        -\frac{1}{U}\sum_{i=1}^U \Bigg( 
            \sum_{k=1}^{X_i} l\left( \boldsymbol{w}^{i}_{\epsilon_i\left(t\right)}\left(t\right), \boldsymbol{x}_{i,k}, y_{i,k} \right) \\
            &+ \mathds{1}_{\left\{ \sum_{b=1}^B a_{i,b}\left(t\right) > 0 \right\}}
            \gamma \left( \frac{1}{X_i \sigma_i} \right)^2 
        \Bigg).
    \end{aligned}
    \end{equation}
    Otherwise, if $M$ users do not satisfy any of these constraints, they will be penalized by $-M^2 \times 10^3$. Here, $- \left(\sum_{i=1}^U \Phi_i \right)^2 \times 10^3$ serves as a strong penalty to enforce constraints (\ref{eq:constraint_a}) to (\ref{eq:constraint_c}). By scaling quadratically, it significantly increases the penalty for multiple violations, which helps speed up training and convergence.
    In addition, $R\left(\boldsymbol{a}\left(t\right)|\boldsymbol{s}\left(t\right)\right)$ increases as the objective function in (\ref{eq:problem_formulation}) decreases, which implies that maximizing the total reward of BSs can minimize the weighted sum of the total loss and the privacy leakage in (\ref{eq:problem_formulation}).
    
    \textcolor{black}{However, we cannot directly obtain the reward $R\left(\boldsymbol{a}\left(t\right)|\boldsymbol{s}\left(t\right)\right)$ based on state $\boldsymbol{s}\left(t\right)$ and action $\boldsymbol{a}\left(t\right)$, since the action $\boldsymbol{a}\left(t\right)$ only determines user association but does not determine the RB allocation strategy and the privacy parameter $\sigma_i$. Hence, in order to obtain the reward $R\left(\boldsymbol{a}\left(t\right)|\boldsymbol{s}\left(t\right)\right)$, we must determine RB allocation vector $\boldsymbol{r}_{i,b}\left(t\right)$ and DP parameter $\sigma_i$ given action $\boldsymbol{a}\left(t\right)$ and state $\boldsymbol{s}\left(t\right)$. In particular, given action  $\boldsymbol{a}\left(t\right)$, problem (\ref{eq:problem_formulation}) can be rewritten as
    \begin{equation} \label{eq:problem_converted}
    \begin{aligned}
    \min_{\substack{\boldsymbol{r}_{i,b}\left(t\right), \\ \sigma_i}} \quad & \frac{1}{\sum_{i=1}^{U}X_i}\left(\sum\limits_{i=1}^U\sum\limits_{b=1}^B\left(a_{i,b}\left(t\right)-\sum\limits_{n=1}^Rr_{i,b}^n\left(t\right) \right)X_i \right.\\
    &\left.\quad + \gamma \sum_{i=1}^{U} \left( \frac{1}{X_i\sigma_i} \right)^2 \sum\limits_{n=1}^R\sum_{b=1}^{B} r_{i,b}^n\left(t\right)\right),
    \end{aligned}
\end{equation}
\[
    \text{s.t.} \quad \textrm{(\ref{eq:constraint_a}), (\ref{eq:constraint_d})$\ -$ (\ref{eq:constraint_f})}.
\]
    % The problem in (\ref{eq:problem_converted}) is a convex optimization problem, and hence the optimal value of $\sigma_i$ can be obtained by standard optimization methods \cite{tavangaran2024differential}.
    % Given action $\boldsymbol{a}\left(t\right)$ and corresponding optimal $\sigma_i$, we can obtain the reward $R(\boldsymbol{a}\left(t\right)|\boldsymbol{s}\left(t\right))$.
    In (\ref{eq:problem_converted}), the first term $\frac{1}{\sum_{i=1}^{U}X_i}\sum\limits_{i=1}^U\sum\limits_{b=1}^B\left(\!a_{i,b}\left(t\right)\!-\!\sum\limits_{n=1}^Rr_{i,b}^n\left(t\right) \!\right)X_i$ is designed to encourage the RB allocation scheme to meet the action $\boldsymbol{a}\left(t\right)$ selected by the BSs. Meanwhile, the first term also encourges the users with more data samples to participate in the CFL training, which is illustrated by our convergence analysis in Theorem 1.
    To solve problem (\ref{eq:problem_converted}), we first optimize $r_{i,b}^n\left(t\right)$ for given $a_{i,b}\left(t\right)$. Then, we optimize $\sigma_i$ for given $r_{i,b}\left(t\right)$. In particular, when $a_{i,b}\left(t\right)$ is given, the optimization problem with respect to $r_{i,b}^n\left(t\right)$ is given by
    \begin{equation} \label{eq:problem_converted_1}
    \begin{aligned}
    \min_{\substack{\boldsymbol{r}_{i,b}\left(t\right)}} \quad & \frac{1}{\sum_{i=1}^{U}X_i}\sum\limits_{i=1}^U\sum\limits_{b=1}^B\left(a_{i,b}\left(t\right)-\sum\limits_{n=1}^Rr_{i,b}^n\left(t\right) \right)X_i,
    \end{aligned}
\end{equation}
\[
    \text{s.t.} \quad \textrm{(\ref{eq:constraint_a}), (\ref{eq:constraint_d})}.
\]
    Since problem (\ref{eq:problem_converted_1}) can be considered as a binary matching problem, it can be directly solved by a standard Hungarian algorithm \cite{wang2022meta}. Next, we optimize $\sigma_i$ for given $r_{i,b}^n\left(t\right)$. In particular, the standard optimization problem is
    \begin{equation} \label{eq:problem_converted_2}
    \begin{aligned}
    \min_{\substack{\sigma_i}}
    \quad \frac{1}{\sum_{i=1}^{U}X_i} \sum_{i=1}^{U}  \gamma \left( \frac{1}{X_i\sigma_i} \right)^2 \sum\limits_{n=1}^R\sum_{b=1}^{B} r_{i,b}^n\left(t\right),
    \end{aligned}
\end{equation}
\[
    \text{s.t.} \quad \textrm{(\ref{eq:constraint_a}), (\ref{eq:constraint_e}), (\ref{eq:constraint_f})}.
\]}
    
    \emph{5) Individual Q function:} The individual Q function $Q_{\boldsymbol{\theta}_b}\left(\boldsymbol{s}_b\left(t\right), \boldsymbol{a}_b\left(t\right)\right)$ is used to estimate the expected future reward under a given state $\boldsymbol{s}_b\left(t\right)$ and a selected action $\boldsymbol{a}_b\left(t\right)$.
    $\boldsymbol{s}_b\left(t\right)$ is the input of the Q function, while the output is a vector of the expected rewards under each action.
    A DNN network parameterized by $\boldsymbol{\theta}_b$ is applied to approximate the local Q function.

    \emph{6) Global Q-function:} The global Q function is defined as $Q_{tot}\left(\boldsymbol{s}\left(t\right), \boldsymbol{a}\left(t\right)\right)$, which is used to estimate the total expected reward of all the BSs.
    Since DPVD-MARL is implemented in a distributed manner, each BS can only obtain the local Q function $Q_{\boldsymbol{\theta}_b}\left(\boldsymbol{s}_b\left(t\right), \boldsymbol{a}_b\left(t\right)\right)$ and it cannot directly obtain the global Q function values. To this end, DPVD-MARL uses local Q functions to approximate the global Q function \cite{cui2019multi}, as follows:  
    \begin{equation} \label{eq:Q_total}
        Q_{tot}\left(\boldsymbol{s}\left(t\right), \boldsymbol{a}\left(t\right)\right) = \sum_{b=1}^BQ_{\boldsymbol{\theta}_b}\left(\boldsymbol{s}_b\left(t\right), \boldsymbol{a}_b\left(t\right)\right).
    \end{equation}
From (\ref{eq:Q_total}), we see that if one BS wants to obtain the global Q function $Q_{tot}\left(\boldsymbol{s}\left(t\right), \boldsymbol{a}\left(t\right)\right)$, it only needs to know the local Q functions of other BSs. 

\subsection{Training of DPVD-MARL}
Next, we introduce how the BSs jointly perform the DPVD-MARL algorithm to solve the problem in (\ref{eq:problem_formulation}).
In particular, we first introduce the loss function used to train the DPVD-MARL models of BSs. 
Then, we explain the training procedure.
% Firstly, each BS $b$ selects an action $\boldsymbol{s}_b\left(t\right)$ based on the current state $\boldsymbol{s}_b\left(t\right)$. 
% Then, each BS $b$ calculates its reward $R_b(\boldsymbol{a}_b\left(t\right)|\boldsymbol{s}_b\left(t\right))$ and share it with other BSs so as to obtain $R(\boldsymbol{a}\left(t\right)|\boldsymbol{s}\left(t\right)) = \sum_{b=1}^BR_{b}(\boldsymbol{a}_b\left(t\right)|\boldsymbol{s}_b\left(t\right))$.
% Given the $\boldsymbol{s}_b\left(t\right)$,$\boldsymbol{r}_b\left(t\right)$ and $\boldsymbol{a}_b\left(t\right)$, each BS $b$ must update its local Q function parameter $\boldsymbol{\theta}_b$.
% However, each BS does not know the global Q function $Q_{tot}(\boldsymbol{s}\left(t\right), \boldsymbol{a}\left(t\right))$ since each BS only has local state $\boldsymbol{s}_b\left(t\right)$. Therefore, it is necessary to utilize Eq.(\ref{eq:Q_total}) to approximate the global Q function $Q_{tot}(\boldsymbol{s}\left(t\right), \boldsymbol{a}\left(t\right))$.
The loss function of DPVD-MARL is
\begin{equation}\label{eq:loss_Q}
\begin{aligned}
    J_t\!\left(\boldsymbol{\theta}_1, \ldots, \boldsymbol{\theta}_B\right) &= \mathbb{E}_{\!\left(\boldsymbol{s}\left(t\right), \boldsymbol{a}\left(t\right), \boldsymbol{s}', \boldsymbol{a}'\right)}\!\Big[Q_{tot}\!\left(\boldsymbol{s}\left(t\right), \boldsymbol{a}\left(t\right)\right) \\
    &\quad - R\!\left(\boldsymbol{a}\left(t\right)\middle|\boldsymbol{s}\left(t\right)\right) - \max_{\boldsymbol{a}'} Q_{tot}\!\left(\boldsymbol{s}', \boldsymbol{a}'\right)\Big]^2,
\end{aligned}
\end{equation}
where $\max_{\boldsymbol{a}'} Q_{tot}\left(\boldsymbol{s}', \boldsymbol{a}'\right)$ is the maximum global Q value at next state $\boldsymbol{s}'$.
The update of local Q function of BS $b$ is
\begin{equation}\label{eq:update_theta}
    \boldsymbol{\theta}_b \gets \boldsymbol{\theta}_b - \lambda_{\boldsymbol{\theta}_b} \nabla_{\boldsymbol{\theta}_b}J_t\left(\boldsymbol{\theta}_1, \ldots, \boldsymbol{\theta}_B\right),
\end{equation}
where $\lambda_{\boldsymbol{\theta_b}}$ is the update rate and $\nabla_{\boldsymbol{\theta_b}}J_t\left(\boldsymbol{\theta}_1, \ldots, \boldsymbol{\theta}_B\right)$ is the gradient of global Q value function, given by
\begin{equation} \label{eq:gradient}
    \begin{aligned}
        &\nabla_{\boldsymbol{\theta_b}}J_t\left(\boldsymbol{\theta}_1, \ldots, \boldsymbol{\theta}_B\right)\\
        &=\nabla_{\boldsymbol{\theta}_b}\!\!\left[Q_{tot}\left(\boldsymbol{s}\!\left(t\right)\!,\! \boldsymbol{a}\!\left(t\right)\right)\!-\! R\left(\boldsymbol{a}\left(t\right)|\boldsymbol{s}\left(t\right)\right) \!-\! \max_{\boldsymbol{a}'} Q_{tot}\left(\boldsymbol{s}'\!,\! \boldsymbol{a}'\right)\right]^2\!\!.
    \end{aligned}
\end{equation}
From (\ref{eq:gradient}), we see that, to update the Q function model of BS $b$, the BS must know $Q_{tot}\left(\boldsymbol{s}\left(t\right), \boldsymbol{a}\left(t\right)\right)$, and thus the Q function model of BS $b$ cannot be updated using local state, action, and reward. To this end, we use (\ref{eq:Q_total}) to approximate $Q_{tot}\left(\boldsymbol{s}\left(t\right), \boldsymbol{a}\left(t\right)\right)$ such that (\ref{eq:gradient}) is %can be
rewritten as 
% \begin{equation} \label{eq:rewritten_gradient}
% \begin{aligned}
%     &\nabla_{\boldsymbol{\theta_b}}J_t(\boldsymbol{\theta}_1, \ldots, \boldsymbol{\theta}_B) \\
%     &= \nabla_{\boldsymbol{\theta}_b}\left[\sum_{b=1}^BQ_{\boldsymbol{\theta}_b}(\boldsymbol{s}_b\left(t\right), \boldsymbol{a}_b\left(t\right))- R(\boldsymbol{a}\left(t\right)|\boldsymbol{s}\left(t\right)) - \max_{\boldsymbol{a}'} Q_{tot}(\boldsymbol{s}', \boldsymbol{a}')\right]^2 \\
%     &= 2 \Delta Q_{tot}(\boldsymbol{s}\left(t\right), \boldsymbol{a}\left(t\right))\nabla_{\boldsymbol{\theta}_b}Q_{b}(\boldsymbol{s}_b\left(t\right), \boldsymbol{a}_b\left(t\right)),
% \end{aligned}
% \end{equation}
\begin{equation} \label{eq:rewritten_gradient}
\begin{aligned}
    &\nabla_{\boldsymbol{\theta_b}}J_t\left(\boldsymbol{\theta}_1, \ldots, \boldsymbol{\theta}_B\right) \\
    &= \!\nabla_{\boldsymbol{\theta}_b}\!\!\left[\!\sum_{b=1}^B\!\!Q_{\boldsymbol{\theta}_b}\!\left(\boldsymbol{s}_b\!\left(t\right), \boldsymbol{a}_b\!\left(t\right)\right)\!-\!\!\!R\!\left(\boldsymbol{a}\!\left(t\right)|\boldsymbol{s}\!\left(t\right)\right)\!\!-\!\!\max_{\boldsymbol{a}'} Q_{tot}\left(\boldsymbol{s}'\!,\! \boldsymbol{a}'\right)\!\right]^2\ \\
    &= 2 \Delta Q_{tot}\left(\boldsymbol{s}\left(t\right), \boldsymbol{a}\left(t\right)\right)\nabla_{\boldsymbol{\theta}_b}Q_{b}\left(\boldsymbol{s}_b\left(t\right), \boldsymbol{a}_b\left(t\right)\right),
\end{aligned}
\end{equation}
where $\Delta Q_{tot}\left(\boldsymbol{s}\left(t\right), \boldsymbol{a}\left(t\right)\right) = \sum_{b=1}^BQ_{\boldsymbol{\theta}_b}\left(\boldsymbol{s}_b\left(t\right), \boldsymbol{a}_b\left(t\right)\right)- R\left(\boldsymbol{a}\left(t\right)|\boldsymbol{s}\left(t\right)\right) - \max_{\boldsymbol{a}'} Q_{tot}\left(\boldsymbol{s}', \boldsymbol{a}'\right)$. 
From (\ref{eq:rewritten_gradient}), we see that each BS only needs the local Q values of other BSs to train its local Q function, thus training MARL models in a distributed manner. 
The specific training process of the DPVD-MARL algorithm is summarized in Algorithm \ref{algo:DPVD-MARL}.
\addtolength{\topmargin}{0.05in}

% \begin{algorithm}[t]
%     \small
%     \caption{DPVD-MARL algorithm for solving problem (\ref{eq:problem_formulation})}
%     \begin{algorithmic}[1]\label{algo:DPVD-MARL}
%         \STATE \textbf{Initialization}: Local model parameter $\boldsymbol{\theta}_b$, learning rate $\lambda_{\boldsymbol{\theta_b}}$, and number of iterations $T$.

%         \FOR{$t = 1, 2, \hdots, T$}
%         \FOR{each environment step}
%         \FOR{each BS}
%             \STATE Record observation of partial environment state $\boldsymbol{s}_b\left(t\right)$.
%             \STATE Select an action according to a $\epsilon$-greedy scheme.
%             \STATE Transmit the $\boldsymbol{a}_b\left(t\right)$ and $\boldsymbol{s}_b\left(t\right)$ to users.
%             \STATE Calculate local rewards.
%         \ENDFOR
%         \FOR{each gradient step}
%             \STATE Calculate the total reward $\boldsymbol{R}(\boldsymbol{a}\left(t\right)|\boldsymbol{s}_b\left(t\right))$ and global Q value function $Q_{tot}(\boldsymbol{s}\left(t\right), \boldsymbol{a}\left(t\right))$.
%             \STATE Update $\{\boldsymbol{\theta}_1, \ldots, \boldsymbol{\theta}_B\}$ by (\ref{eq:update_theta})
%         \ENDFOR
%         \ENDFOR
%         \ENDFOR

%     \end{algorithmic}
% \end{algorithm}
\setlength{\textfloatsep}{8pt}
\begin{algorithm}[t]
    \normalsize
    \caption{DPVD-MARL Algorithm for Solving Problem (\ref{eq:problem_formulation})}
    \begin{algorithmic}[1]\label{algo:DPVD-MARL}
        \STATE \textbf{Initialization}: Local model parameters $\left\{\boldsymbol{\theta}_1, \ldots, \boldsymbol{\theta}_B\right\}$, learning rate $\lambda_{\boldsymbol{\theta}_b}$, and number of iterations $T$;
        \FOR{each epoch}
        \FOR{each episode} 
        \FOR{each step $t = 1, 2, \hdots, T$}
        \FOR{each BS $b = 1, 2, \ldots, B$}
            \STATE Observe partial environment state $\boldsymbol{s}_b\left(t\right)$;
            \STATE Select an action $\boldsymbol{a}_b\left(t\right)$ using an $\epsilon$-greedy policy;
            \STATE Calculate local rewards;
            \STATE Transmit $\boldsymbol{a}_b\left(t\right)$ and local rewards to other BSs;
            \STATE Calculate the total reward $\boldsymbol{R}\left(\boldsymbol{a}\left(t\right)|\boldsymbol{s}\left(t\right)\right)$ and global Q-value function $Q_{tot}\left(\boldsymbol{s}\left(t\right), \boldsymbol{a}\left(t\right)\right)$;
        \ENDFOR
        \ENDFOR
        \ENDFOR
        \STATE Each BS $b$ updates $\boldsymbol{\theta}_b$ based on (\ref{eq:update_theta});
        \ENDFOR

    \end{algorithmic}
\end{algorithm}

\subsection{Convergence, Implementation, and Complexity Analysis}

\textcolor{black}{Next, we analyze the convergence, implementation, communication overhead, and complexity of the proposed DPVD-MARL method.}

\emph{1) Convergence analysis:} Here, we analyze the convergence of the proposed DPVD-MARL algorithm. 
Let $\boldsymbol{\pi}_b\left(\boldsymbol{s}_b\left(t\right), \boldsymbol{a}_b\left(t\right)\right)$ be the conditional probability of BS $b$ choosing action $\boldsymbol{a}_b\left(t\right)$ given its current state $\boldsymbol{s}_b\left(t\right)$. 
Define the gap between the actual global Q function $Q^{\pi}\left(\boldsymbol{s}\left(t\right), \boldsymbol{a}\left(t\right)\right)$ and the global Q function $Q^{\pi}_{tot}\left(\boldsymbol{s}\left(t\right), \boldsymbol{a}\left(t\right)\right)$ approximated by Q networks under a policy $\pi$ as
\begin{equation} \label{eq:omega}
    \Omega^{\pi}\left(\boldsymbol{s}\left(t\right), \boldsymbol{a}\left(t\right)\right) = Q^{\pi}\left(\boldsymbol{s}\left(t\right), \boldsymbol{a}\left(t\right)\right) - Q^{\pi}_{tot}\left(\boldsymbol{s}\left(t\right), \boldsymbol{a}\left(t\right)\right),
\end{equation}
where
\begin{equation} \label{eq:Q_pi}
\begin{aligned}
    Q^{\pi}\left(\boldsymbol{s}\left(t\right), \boldsymbol{a}\left(t\right)\right) &= \sum_{\boldsymbol{s}'_{t+1}}P\left(\boldsymbol{s}'_{t+1} \mid \boldsymbol{s}\left(t\right), \boldsymbol{a}\left(t\right)\right) \\
    & \quad \times\! \left[\! R\left(\boldsymbol{a}\left(t\right) \mid \boldsymbol{s}\left(t\right)\right) \!+ \!\max_{\boldsymbol{a}'_{t+1}} Q^\pi\!\!\left(\boldsymbol{s}'_{t+1}, \boldsymbol{a}'_{t+1}\right) \!\right]\!,
\end{aligned}
\end{equation}
with $\boldsymbol{s}'_{t+1}$ being the global state and $\boldsymbol{a}'_{t+1}$ being the joint action of all BSs at the next time step of iteration $t$, and $P\left(\boldsymbol{s}'_{t+1}|\boldsymbol{s}\left(t\right), \boldsymbol{a}\left(t\right)\right)$ being the transition probability from the current global state $\boldsymbol{s}\left(t\right)$ to the next global state $\boldsymbol{s}'_{t+1}$ given the joint action $\boldsymbol{a}\left(t\right)$.
Given these definitions, the convergence of the proposed DPVD-MARL is summarized in the following lemma. 
\begin{lemma} \label{lem:1}
    If (i) $\varepsilon \to 0$ and (ii) $\left| \Omega^{\pi_1}\left(s, a\right) - \Omega^{\pi_2}\left(s, a\right) \right| \leq \varepsilon$ for any policies $\pi_1, \pi_2$,
our proposed DPVD-MARL method can converge to the optimal $Q^*_{tot}$ \cite{wang2022distributed}.
\end{lemma}
\noindent From Lemma \ref{lem:1}, we see that the convergence of the proposed DPVD-MARL algorithm depends on the gap between the Q function approximated by the neural network and the ground truth Q function. In our algorithm, both conditions are satisfied since the error introduced by the Q function approximation decreases as the neural networks are trained over time, and the value decomposition method used to generate global Q value in (\ref{eq:Q_total}) ensures that the difference between the actual global Q-function and its approximated value remains bounded. Our simulation results in Section V also demonstrates that the proposed DPVD-MARL algorithm effectively converges under the networks with different number of users.

\emph{2) Implementation analysis:} Next, we describe the implementation of the proposed DPVD-MARL method.
The proposed DPVD-MARL method includes a training stage and a decision-making stage.
In the training stage, each BS requires (i) the task identity $\epsilon_i\left(t\right)$ of each user $i$ and (ii) the distance $d_{b,i}\left(t\right)$ between each BS $b$ and user $i$. To get task identity information, the BS needs to collect the task identity vector of all the users at each CFL training iteration. The distance information is estimated by the signals transmitted from each user to the BS. To calculate the global Q functions, the BS needs to collect individual Q functions as shown in (\ref{eq:Q_total}) in our training stage.
In the decision-making stage, the well trained Q value function can be directly used by the BS to determine user association, RB allocation, and DP noise. 

\textcolor{black}{\emph{3) Communication overhead analysis:} Next, we analyze the communication overhead~\cite{xu2023edge, yao2024wireless} of the proposed method.
During an uplink FL model transmission stage, each user uploads one local model $\boldsymbol{w}^i_{\epsilon_i(t)}(t+1)$ to its associated BS. Let $D$ be the size of each local model. From Eq. (7), we see that the uplink transmission delay per user is $\frac{D}{c_{i,b,t}^\mathrm{U}\left(\boldsymbol{r}_{i,b}(t)\right)}$, and the corresponding energy consumption is $P_i \times \frac{D}{c_{i,b,t}^\mathrm{U}\left(\boldsymbol{r}_{i,b}(t)\right)}$.
In a downlink global model transmission stage, each BS broadcasts the global models to all users. Let $P_{\textrm{S}}$ be the transmit power of each BS. The downlink transmission delay from BSs to each user $i$ is $\frac{D}{c^{\textrm{D}}_t}$, and the corresponding energy consumption is $P_{\textrm{S}}\times\frac{D}{c^{\textrm{D}}_t}$.}
\textcolor{black}{Hence, the total communication overhead per iteration is $\sum_{i \in \mathcal{U}}\frac{P_iD}{c_{i,b,t}^\mathrm{U}\left(\boldsymbol{r}_{i,b}(t)\right)}+\sum_{b \in \mathcal{B}}\frac{P_{\textrm{S}}D}{c^{\textrm{D}}_t}$. Here, the uplink communication overhead has been optimized, and the down link communication overhead is limited by $\mathcal{K}_b$. To further reduce the downlink communication overhead, the BSs can also implement our designed RL method to determine the tasks that each BS should broadcast.}

% \textcolor{red}{In this work, we also consider the scenario where the system already knows the number of data distributions \( K \). If \( K \) changes over time, the proposed DPVD-MARL can still serve as the lower layer of a hierarchical RL (HRL) framework, where the upper layer uses the CFL clustering accuracy as its state and selects \( K \) as the action to adjust the number of task identities.}

\textcolor{black}{\emph{3) Complexity analysis:}
The computational complexity of the proposed DPVD-MARL algorithm consists of three main components: (i) the DPVD-MARL training process, (ii) the optimization of DP noise parameters, and (iii) the RB allocation optimization using the Hungarian algorithm.}

\textcolor{black}{\textbf{1) DPVD-MARL Training Complexity:} Each BS trains a deep neural network (DNN) to approximate its local Q-function $Q_{\boldsymbol{\theta}_b}\left(\boldsymbol{s}_b\left(t\right), \boldsymbol{a}_b\left(t\right)\right)$. The training complexity depends on the architecture of the DNN, specifically the number of layers and neurons. In each iteration, the computation of the data or gradient transmission in the DNN has a complexity of $\mathcal{O}\left(N_b\right)$, where $N_b$ is the number of parameters in the DNN of BS $b$.}

\textcolor{black}{\textbf{2) RB Allocation Complexity:} The RB allocation problem, formulated as a binary matching problem, is solved using the Hungarian algorithm. For a matching matrix of size $n \times n$, the algorithm has a complexity of $\mathcal{O}\left(n^3\right)$. Here, our matching matrix depends on the number of users (i.e., $U$) and the number of RBs (i.e., $R$). Since $U \geq R$, the complexity per BS is $\mathcal{O}\left(U^3\right)$. For $B$ BSs, the total complexity is $\mathcal{O}\left(B \times U^3\right)$.
This complexity of the matching theory based Hungarian algorithm is lower compared to other optimization methods (e.g., genetic algorithm (GA) in \cite{deng2024task}) used to solve integer programming problems. }

\textcolor{black}{\textbf{3) DP Noise Optimization Complexity:} The DP noise parameters $\sigma_i$ are optimized by the interior-point method \cite{nuhoglu2024source}. Hence, the computational complexity per iteration is $\mathcal{O}(U^3)$ \cite{boyd2004convex}.
The number of iterations required for convergence of the interior-point method is $\mathcal{O}\left(\sqrt{U} \log\left(1/\beta\right)\right)$, where $\beta$ is the desired solution accuracy. Therefore, the total computational complexity for solving the DP noise optimization problem is}

\textcolor{black}{\begin{equation}
\mathcal{O}\left(U^3\right) \times \mathcal{O}\left(\sqrt{U} \log\left(\frac{1}{\beta}\right)\right) = \mathcal{O}\left(U^{3.5} \log\left(\frac{1}{\beta}\right)\right).
\end{equation}}

\textcolor{black}{Given the component-wise complexity analysis, the overall computational complexity of the entire algorithm is
\begin{equation}
\mathcal{O}\left( N + U^{3.5} \log\left(1/\beta\right) + B \times U^3 \right),
\end{equation}
where $N = \max\left(N_1, N_2, \dots, N_B\right)$. }

\section{Simulation Results and Analysis}

In our simulations, we consider a network consisting of $B = 4$ BSs and $U = 12$ users, distributed randomly within a $\left(500 \times 500\right)$$m^2$ area. 
% Each BS has a coverage radius of $500$ meters. 
 Other simulation parameters are provided in Table~\ref{tb:simu_para} \cite{tavangaran2024differential}.
We use the CIFAR-10 dataset \cite{sattler2020clustered} for training the FL model that consists of two convolutional layers with $3 \times 3$ convolutional kernels, and three fully-connected layers ($64 \times 8 \times 8 \times 128$, $128 \times 64$, and $64 \times 10$).
% As the example in Fig. \ref{fig:simulation_5} shows, we assign users non-IID data distributions by varying proportions of data types and dataset sizes, ensuring users in the same cluster share dominant data characteristics while maintaining inter-cluster heterogeneity.
% For comparison purpose, we consider baselines: a) the independent Q-learning (IQL) in \cite{wang2019distributed} and monotonic value function factorisation (QMIX) in \cite{rashid2020monotonic}, b) the traditional penalty scheme, c) the traditional block coordinate descent (BCD) optimization scheme in \cite{wright2006numerical}, and d) the traditional CFL clustering method in \cite{ghosh2022efficient}.
\textcolor{black}{For comparison purposes, we consider three baselines:
a) the independent Q-learning (IQL) \cite{wang2019distributed} in which each agent (BS) independently learns its own user selection strategy without the collaborations with other BSs.
b) The monotonic value function factorization (QMIX) in \cite{rashid2020monotonic}, which employs a neural network to aggregate local Q values of all BSs to approximate the global Q value function. Both methods only determine the user selection vector of each BS (i.e., $\boldsymbol{a}_b\left(t\right)$). The optimal DP noise (e.g., $\sigma_i$) and RB allocation strategies (i.e., $r_{i,b}^n\left(t\right)$) are optimized by the proposed optimization method in this paper. Hence, baselines a) and b) are used to show the impact of different user selection strategies on the weighted sum of the CFL convergence and DP noise.
c) Standard CFL method in \cite{ghosh2022efficient}, where the cluster identity of each
device only considers the training loss (i.e., $\lambda = 0$ in (\ref{eq:identity})). Baseline c) is used to show the benefits of our proposed device task identity determination method.}

% c) **Traditional Penalty Scheme**: This baseline incorporates fixed, large penalties for invalid actions during the learning process. Although straightforward to implement, such schemes can discourage exploration and lead to slower convergence or suboptimal policies, particularly in dynamic environments where valid actions are sparsely distributed.

% d) **Traditional Block Coordinate Descent (BCD) Optimization Scheme**: This method, based on \cite{wright2006numerical}, iteratively optimizes one set of variables while keeping others fixed. While effective for certain types of optimization problems, BCD can be computationally expensive and slow to converge when applied to large-scale problems involving multiple decision variables, such as resource allocation and user clustering in CFL.

\begin{table}[t]
\captionsetup{labelfont={color=blue},textfont={color=blue}}
\caption{Simulation Parameters \cite{dwork2006calibrating, dwork2014algorithmic}}
\label{tb:simu_para}
\centering
\fontsize{9}{9}\selectfont{
{\color{blue}\begin{tabular}{|c|c||c|c|}
\hline
\textbf{Parameters}         & \textbf{Value}  & \textbf{Parameters}         & \textbf{Value}  \\ \hline
$\gamma$ ($\times 10^{7}$) & 2 & $V_{\textrm{max}}$ & 12 \\ \hline
$K$ & 4 & $N_{\textrm{min}}$ & 100 \\ \hline
$R$ & 4 & $\gamma_T$ (Mbps) & 1 \\ \hline
$B^{\textrm{U}}$ (MHz) & 1 & $\gamma_D$ (ms) & 10 \\ \hline
$W_0$ (dB) & -90 & $P_i$ (dBm) & 30 \\ \hline
$\alpha$ & 0.03 & $B$ & 4 \\ \hline
$\lambda_{\theta_b}$ & 0.0015 & $\zeta_2$ & 0.05-0.2 \\ \hline
\end{tabular}}
}
\end{table}

\begin{figure}[t]
    \centering
\includegraphics[width=.48\textwidth]{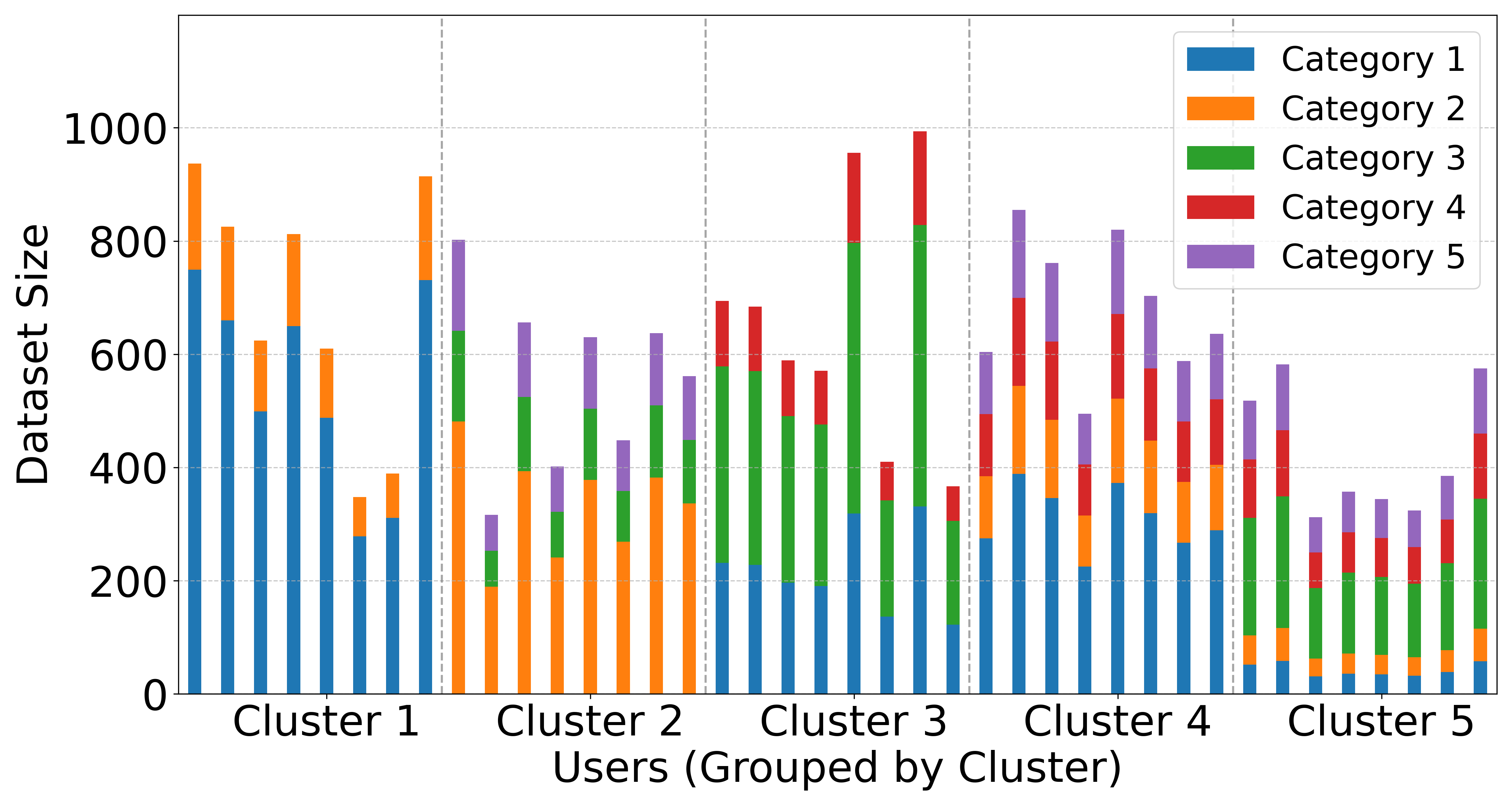}
    \caption{Example of cluster-wise users with non-IID data.}
    \label{fig:simulation_5}
    %\vspace{-0.1cm}
\end{figure}

\begin{figure}[tp]
    \centering
\includegraphics[width=.45\textwidth]{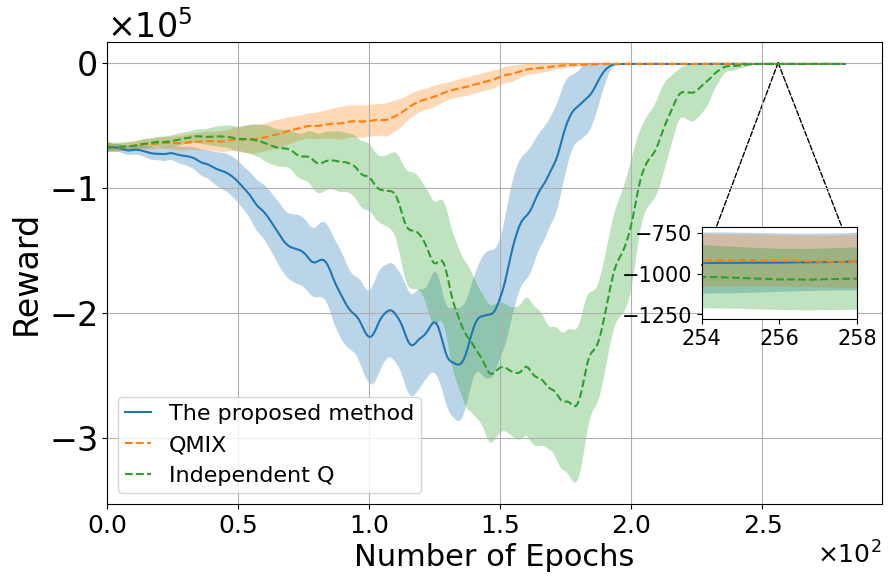}
    \vspace{-0.2cm}
    \caption{Convergence of the proposed DPVD-MARL, independent Q, and QMIX.}
    \label{fig:simulation_1}
    \vspace{-0.3cm}
\end{figure}

Fig. \ref{fig:simulation_5} shows the data distribution of the users in various clusters. In this figure, different colors represent different data categories. For example, red color represents the category of cats, while green color represents the category of trucks.
From this figure, we see that the size of the dataset of the users in each cluster are different, which reflects the heterogeneous nature of the non-IID data distribution. 
We also see that the users in a cluster have the same categories of data and similar data distribution. 
However, the users in different clusters have different categories of data or different data distribution. 
For example, in this figure, users in cluster 1 possess data from categories 1 and 2, with a distribution 4:1. In contrast, users in cluster 2 have data from categories 2, 3, and 5, with a distribution 3:1:1. 
% This distinct data distribution within clusters serves as the basis for distinguishing different clusters.

In Fig. \ref{fig:simulation_1}, we show the convergence of DPVD-MARL, IQL, and QMIX during the training process. \textcolor{black}{In this figure, the temporary performance drop is due to the invalid actions that violate communication and delay constraints thus introducing a large penalty in the reward.} From this figure, we also see that, compared with IQL, our proposed method can improve the convergence rate by up to 20\%, and the accumulated reward by 15\% when the number of epochs is over 260. 
This is because the proposed DPVD-MARL method uses (\ref{eq:Q_total}) to approximate $Q_{tot}\left(\boldsymbol{s}\left(t\right), \boldsymbol{a}\left(t\right)\right)$, allowing BSs to collaboratively find a globally optimal solution.
We also see that our proposed method can achieve similar performance in terms of convergence speed and accumulated reward as the QMIX method. This implies that our proposed method is more efficient since our proposed method directly aggregates all local Q values to approximate global Q value instead of using neural networks as implemented in QMIX.

\begin{figure}[tp]
    \centering
\includegraphics[width=.43\textwidth]{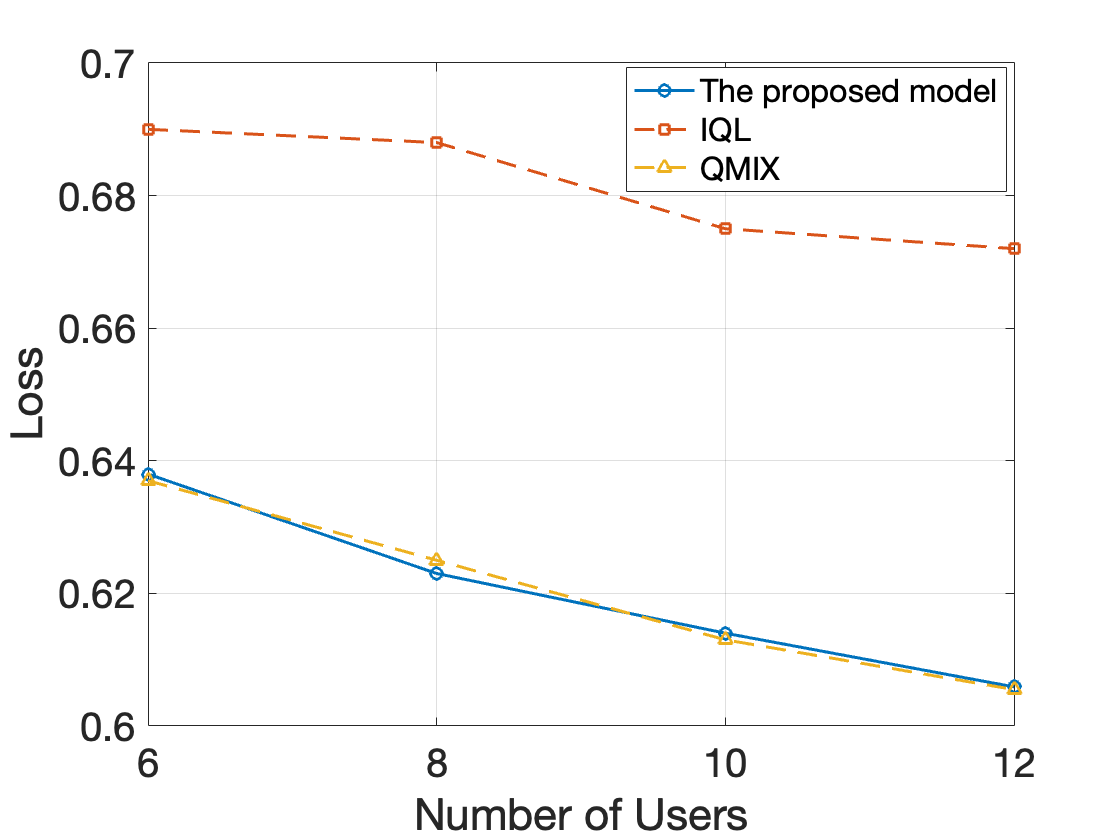}
    \caption{ CFL convergence loss as the number of users varies. }
    \label{fig:simulation_6}
    \vspace{-0.3cm}
\end{figure}

In Fig. \ref{fig:simulation_6}, we show how the CFL convergence of DPVD-MARL, IQL and QMIX varies when the number of users changes. From this figure, we see that, compared to IQL, our proposed method can reduce the CFL loss by up to 10\% when the network has $12$ users. 
This is because DPVD-MARL enables effective collaboration among BSs, allowing them to coordinate their actions to include more users for CFL training. 
Fig. \ref{fig:simulation_6} also shows that the loss resulting from our proposed method decreases faster than IQL as the number of users increases. This is because our proposed method enables the BSs to efficiently select the users for CFL training and allocate appropriate RBs to the users. 
Fig. \ref{fig:simulation_6} also shows that the proposed method can achieve a similar performance as QMIX. This is due to the fact that our team reward is designed as the summation of the individual BS rewards. As a result, directly aggregating local Q-values to approximate the global Q-value as done in (\ref{eq:Q_total}) provides a reasonably accurate estimation.

\begin{figure}[t]
    \centering
\includegraphics[width=.45\textwidth]{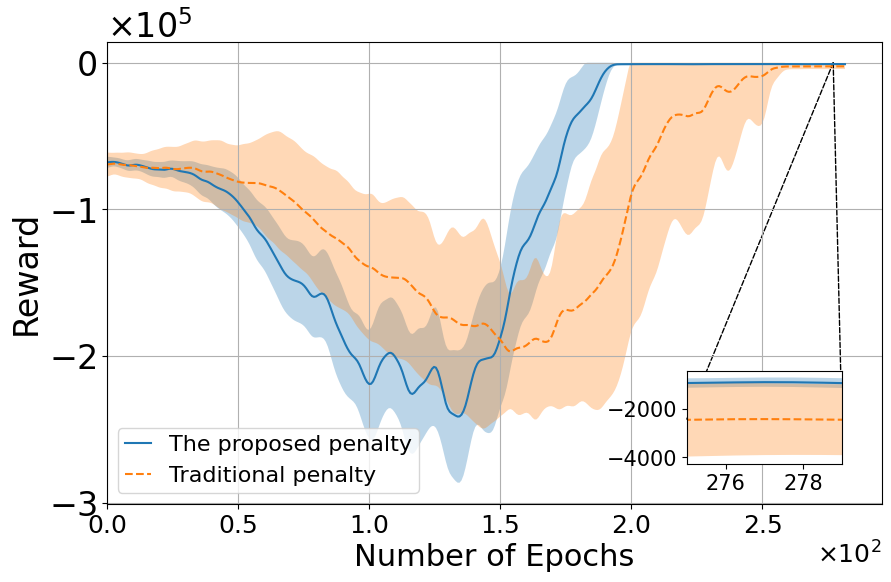}
    \caption{ Impact of penalty assignment schemes on CFL convergence. }
    \label{fig:simulation_2}
   \vspace{-0.3cm}
\end{figure}

In Fig. \ref{fig:simulation_2}, we show how the penalty assignment scheme affects DPVD-MARL training. In this figure, we compare our proposed method with the standard DPVD-MARL method that assigns a single, large penalty ($-5 \times 10^3$) to invalid actions. From this figure, we see that our scheme can improve convergence rate by up to 25\%, and reduce the variance (narrower shaded area) by at least 80\% after 260 epochs. This is due to the fact that the proposed penalty approach allows the proposed method to learn the gap between the valid and invalid actions, thus guiding it to avoid invalid actions and select only valid actions.
We also see that the convergence value of the standard method is approximately 1500 lower than that of our proposed method. This is because the excessively large penalty makes the model overly sensitive to invalid actions, which increases the risk of converging to a local minimum.

\begin{figure}[tp]
    \centering
\includegraphics[width=.45\textwidth]{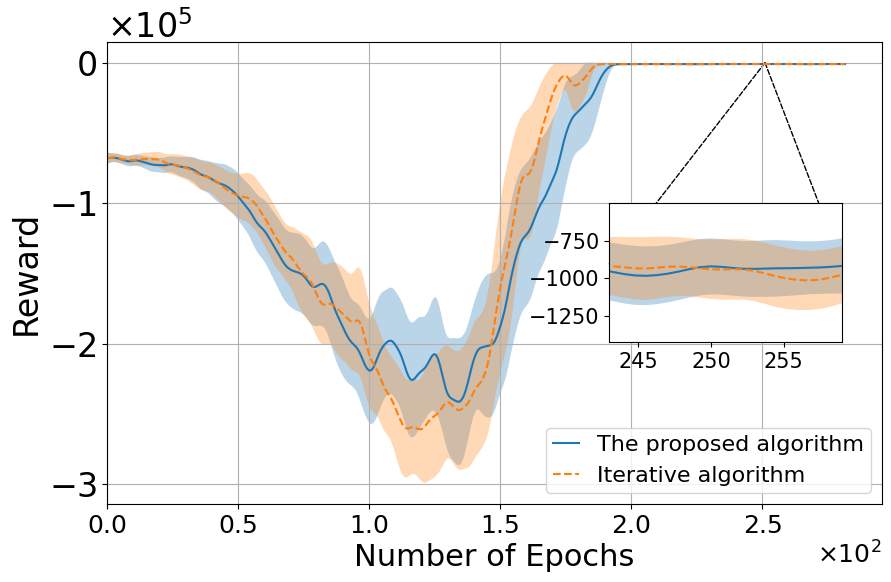}
    \caption{ The impact of the RB allocation and noise optimization schemes on CFL convergence. }
    \label{fig:simulation_3}
   \vspace{-0.2cm}
\end{figure}

Fig. \ref{fig:simulation_3} shows the impact of optimization schemes on CFL convergence. In this figure, we compare our proposed method that first optimizes RB allocation strategy $r_{i,b}^n\left(t\right)$ and then optimizes DP noise $\sigma_i$ without iterative processes to the standard block coordinate descent (BCD) optimization method \cite{wright2006numerical} that solves the problem in (\ref{eq:problem_converted}) by iteratively optimizing $r_{i,b}^n\left(t\right)$ and $\sigma_i$. 
From this figure, we see that our proposed optimization method with lower computational complexity (i.e., $\mathcal{O}\left(U^{3.5} \log\left(1/\beta\right) + B \times U^3 \right)$) can achieve the same performance as the standard iterative optimization method with a computational complexity $\mathcal{O}\left(A\left(U^3 + U^{3.5}\log\left(1/\beta\right)\right)\right)$, where $A$ is the number of iterations needed for BCD convergence \cite{richtarik2014iteration}. 
% Assuming $A$ iterations are needed for BCD convergence, the total complexity is computed as $\mathcal{O}(A(U^3 + U^{3.5}log(1/\epsilon)))$ \cite{beck2014introduction}, which cost more energy than our proposed method, thus increasing the training time with epoch increases. This demonstrates that our proposed model has same effectiveness but higher efficiency.
This is because, although our proposed method introduces initial errors by independently optimizing $r_{i,b}^n\left(t\right)$ and $\sigma_i$, these errors are eliminated during the training process. 
\begin{figure}[t]
    \centering
\includegraphics[width=.49\textwidth]{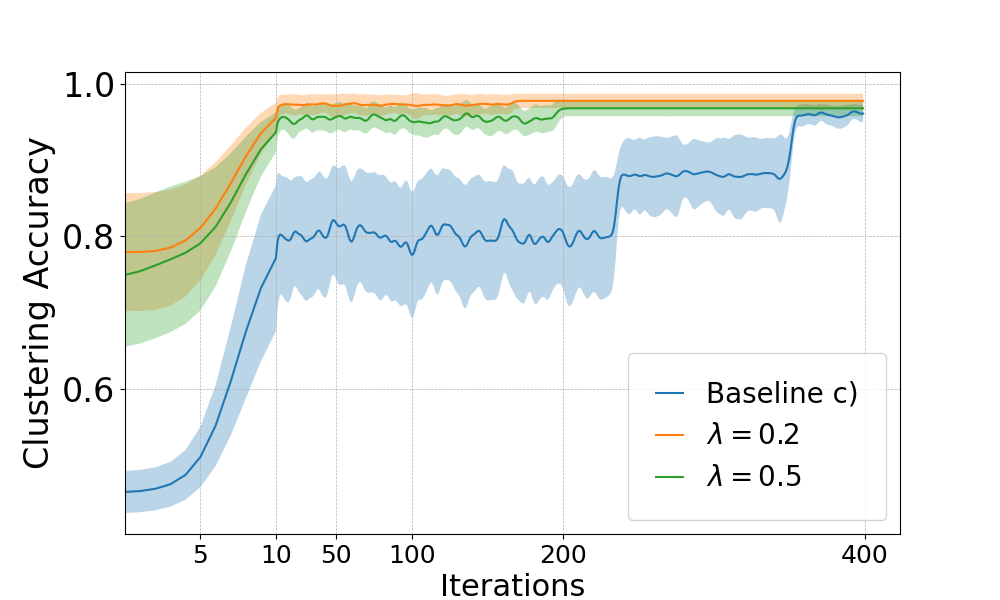}
    \caption{ Clustering accuracies of CFL. }
    \label{fig:simulation_4}
    \vspace{-0.4cm}
\end{figure}

% In Fig. \ref{fig:simulation_4}, we present the clustering accuracy rates of our proposed CFL algorithm with different $\lambda$ values compared to the standard CFL method, which focuses solely on minimizing loss for task identity determination. The results demonstrate that our method achieves higher clustering accuracy and faster convergence, especially for $\lambda = 0.2$, by jointly considering gradient similarity and loss values during task identity determination.
In Fig. \ref{fig:simulation_4}, we show how the clustering accuracies of the considered algorithms change with $\lambda$. Here, clustering accuracy is defined as the proportion of users correctly assigned to their respective clusters at the convergence stage of CFL. From this figure, we see that, our proposed method improves clustering accuracy by up to 5\% compared to the baseline c) that only uses training loss to determine the task identity. This is because the proposed method jointly considers the gradient direction and minimum loss to determine the task identity of each user. 
We also see that the proposed method with $\lambda = 0.2$ outperforms the algorithm with $\lambda = 0.5$ in both convergence speed and final clustering accuracy.
This implies that using gradient direction can improve CFL performance. However, if we increase the weight of gradient direction, the CFL performance may be degraded.
\textcolor{black}{We also see that the clustering accuracy of the proposed method reaches $96\%$ after 15 iterations, which means that the BSs only need to broadcast a subset of task models in the following training iterations, thereby reducing downlink communication overhead.}

\begin{figure}[tp]
    \centering
\includegraphics[width=.48\textwidth]{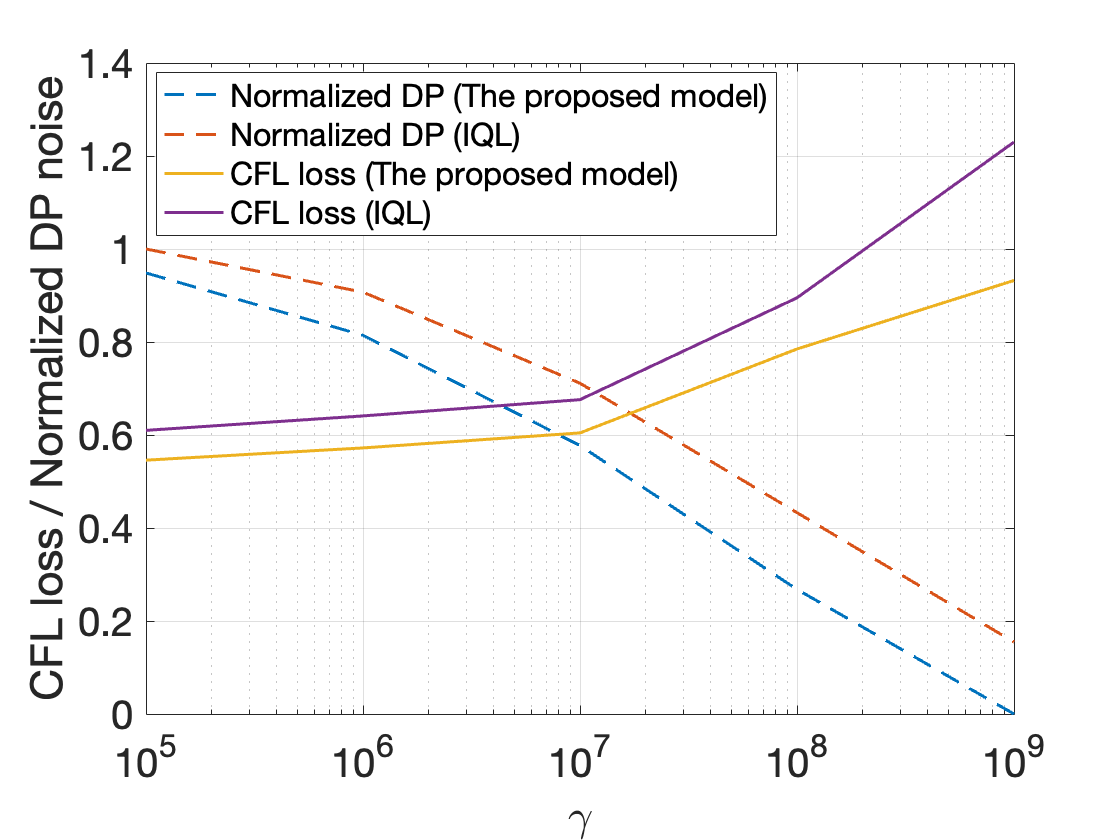}
    \vspace{-0.2cm}
    \caption{ Impact of $\gamma$ on CFL convergence loss and DP noise. }
    \label{fig:simulation_7}
    \vspace{-0.3cm}
\end{figure}

In Fig. \ref{fig:simulation_7}, we show how $\gamma$ in (\ref{eq:problem_formulation}) affects the DP noise and CFL loss. 
From this figure, we see that, as $\gamma$ increases, the DP noise decreases, while the CFL loss increases. This is because when $\gamma$ is small, both considered methods focus more on minimizing the total CFL loss as shown in (\ref{eq:reward_satisfy}). In contrast, when $\gamma$ is large (e.g., $\gamma \geq 10^8$), the considered methods focus more on the DP noise optimization, such that the methods discourage the users to participate in CFL training, thus decreasing the CFL performance. 
  When $\gamma$ is around $10^7$, the proposed method achieves an effective balance between minimizing CFL loss and reducing DP noise, resulting in an optimal performance.
We also see that the CFL loss of the proposed method increases slowly, and its normalized DP noise decreases rapidly compared to IQL as $\gamma$ increases. 
Fig. \ref{fig:simulation_7} also shows that when $\gamma = 10^9$, the proposed method can reduce the training loss and normalized DP noise by up to 24\% and 18\%, respectively. 
% This is because both of the considered models have few RBs to allocate with very large $\gamma$. Hence, with limited user connections, the proposed model jointly determining global actions will amplify the gap of the CFL loss.

\begin{figure}[tp]
\captionsetup{labelfont={color=black},textfont={color=black}}
    \centering
\includegraphics[width=.46\textwidth]{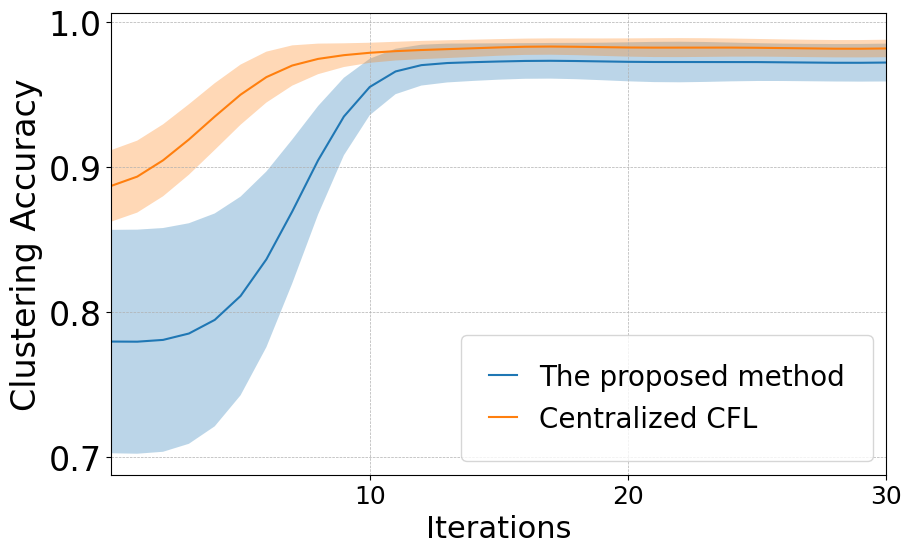}
    \vspace{-0.2cm}
    \caption{ Clustering accuracies of the proposed method and centralized CFL. }
    \label{fig:simulation_8}
    \vspace{-0.3cm}
\end{figure}

\textcolor{black}{Fig.~\ref{fig:simulation_8} shows the clustering accuracy of the proposed method and centralized CFL in which the BSs cluster the local FL models of the users using the K-means method \cite{ghosh2022efficient}. From this figure, we see that the clustering accuracy of the proposed method is $2.2\%$ lower compared to the centralized CFL when the number of iterations is $15$. This is because centralized CFL requires the BSs to check the local models of all users. Meanwhile, we find that, the proposed method requires 14\% more energy for user to calculate loss, update local models, and determine task identities but reduces the training time by up to 26\%. This is because our proposed method enables the users to compute and determine task identities in a distributed manner thus reducing the computational time of model clustering.}

\begin{figure}[tp]
\captionsetup{labelfont={color=black},textfont={color=black}}
    \centering
\includegraphics[width=.46\textwidth]{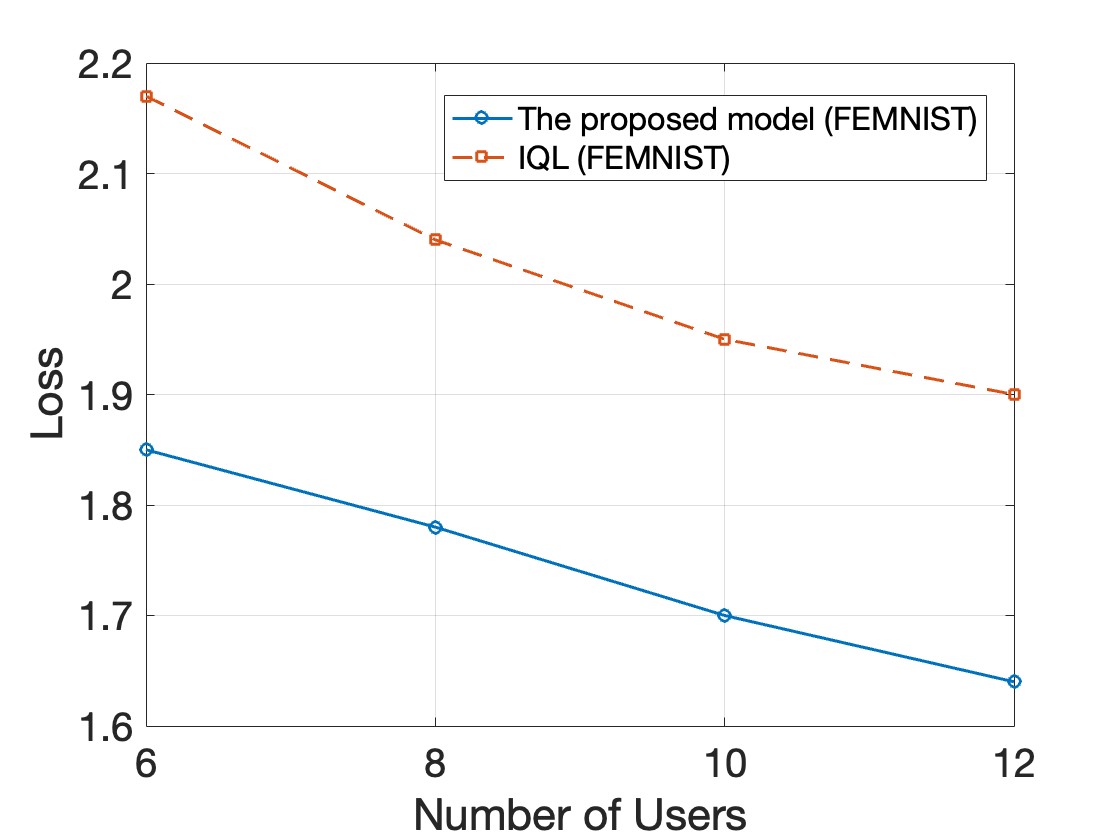}
    \vspace{-0.2cm}
    \caption{ CFL training loss of the proposed method and IQL as the number of users varies using FEMNIST dataset. }
    \label{fig:simulation_9}
    \vspace{-0.3cm}
\end{figure}

\textcolor{blue}{Fig.~\ref{fig:simulation_9} shows how the CFL training loss of the proposed method varies as the number of users increases using the FEMNIST dataset. From the figure, we see that our method reduces the CFL loss by up to $13.7\% $ compared to IQL when the network has $12$ users. This is because the proposed method jointly performs user grouping and local model update across all users, which allows the BSs to better capture task differences thus improving training stability.}

\section{Conclusions}\label{Conclusions}
In this paper, we developed a secure, communication-efficient, and multi-task CFL framework.
In particular, we considered a CFL framework where BSs and users jointly perform CFL training. 
We formulated an optimization problem that aims to minimize the overall CFL training loss by optimizing RB allocation and user scheduling, while considering transmission delays and DP constraints. To solve this problem, we designed a novel DPVD-MARL algorithm, enabling distributed BSs to independently manage user associations, RB allocation, and DP noise levels, while collaboratively optimizing the training performance across all learning tasks. 
To further improve MARL performance, we designed a novel penalty assignment scheme that adjusts penalties based on the number of users that violate the communication constraints, guiding the algorithm to faster convergence. 
Simulation results demonstrated that our proposed %DPVD-MARL 
algorithm outperforms the state-of-the-art methods. 
\textcolor{black}{In future, we will study the use of analog transmission techniques such as over-the-air computation (AirComp) to further improve the FL model transmission efficiency, and extend the proposed CFL framework for scenarios where user data distributions change over time.}

\appendix
\renewcommand{\thesubsection}{}
\subsection*{Proof of Theorem 1}
To prove Theorem 1, we first rewrite $F\left(\boldsymbol{w}_k^G\left(t\right)\right)$ using the second-order Taylor expansion, which is expressed as
\begin{equation} \label{eq:taylor_1}
\begin{split}
    &F \left( \boldsymbol{w}_k^G\left(t+1\right) \right) \\
    &= F \left( \boldsymbol{w}_k^G\left(t\right) \right) 
    + \left( \boldsymbol{w}_k^G\left(t+1\right) - \boldsymbol{w}_k^G\left(t\right) \right)^T \nabla F \left( \boldsymbol{w}_k^G\left(t\right) \right) \\
    &\quad + \frac{1}{2} \left( \boldsymbol{w}_k^G\left(t+1\right) - \boldsymbol{w}_k^G\left(t\right) \right)^T 
    \nabla^2 F \left( \boldsymbol{w}_k^G\left(t\right) \right) \\
    &\quad \times \left( \boldsymbol{w}_k^G\left(t+1\right) - \boldsymbol{w}_k^G\left(t\right) \right),\\
    &\leq F \left( \boldsymbol{w}_k^G\left(t\right) \right) 
    + \left( \boldsymbol{w}_k^G\left(t+1\right) - \boldsymbol{w}_k^G\left(t\right) \right)^T \nabla F \left( \boldsymbol{w}_k^G\left(t\right) \right) \\
    &\quad + \frac{L}{2} \left\| \boldsymbol{w}_k^G\left(t+1\right) - \boldsymbol{w}_k^G\left(t\right) \right\|^2,
\end{split}
\end{equation}
% where the inequality stems from the assumption in (\ref{eq:twice}). 
% Since $....$  (\ref{eq:twice}), we can derive the inequality of $F \left( \boldsymbol{w}_k^G(t+1) \right)$ from (\ref{eq:taylor_1}) that
% \begin{equation} \label{eq:talor}
% \begin{aligned}
%     &F \left( \boldsymbol{w}_k^G(t+1) \right) \\
%     &\leq F \left( \boldsymbol{w}_k^G\left(t\right) \right) 
%     + \left( \boldsymbol{w}_k^G(t+1) - \boldsymbol{w}_k^G\left(t\right) \right)^T \nabla F \left( \boldsymbol{w}_k^G\left(t\right) \right) \\
%     &\quad + \frac{L}{2} \left\| \boldsymbol{w}_k^G(t+1) - \boldsymbol{w}_k^G\left(t\right) \right\|^2.
% \end{aligned}
% \end{equation}
where the inequality stems from the assumption in (18). Given the learning rate $\alpha = \frac{1}{L}$ and (12), the expected value of $F\left(\boldsymbol{w}_k^G\left(t+1\right)\right)$ with respect to $p_{k,i}^t$ is
\begin{equation}\label{eq:transfer}
\begin{split}
    &\mathbb{E} \left( F\left(\boldsymbol{w}_k^G\left(t+1\right)\right) \right) \leq \\ &\mathbb{E} \left( F\left(\boldsymbol{w}_k^G\left(t\right)\right) - \alpha \left(\nabla F\left(\boldsymbol{w}_k^G\left(t\right)\right) - \boldsymbol{o}_k + \boldsymbol{e}_k + \boldsymbol{n}_k\right)^T  \right. \\
    &\left. \quad \times \nabla F\left(\boldsymbol{w}_k^G\left(t\right)\right) + \frac{L \alpha^2}{2} \left\| \nabla \! F\!\left(\boldsymbol{w}_k^G\left(t\right)\right)\! -\! \boldsymbol{o}_k \!+\! \boldsymbol{e}_k \!+\! \boldsymbol{n}_k\right\|^2 \!\right) \\
    &\overset{(a)}{=} \mathbb{E} \left( F\left(\boldsymbol{w}_k^G\left(t\right)\right) \right) - \frac{1}{2L} \left\| \nabla F\left(\boldsymbol{w}_k^G\left(t\right)\right) \right\|^2 \\
    &\quad+ \frac{1}{2L} \mathbb{E} \left( \left\|\boldsymbol{o}_k - \boldsymbol{e}_k + \boldsymbol{n}_k\right\|^2 \right) \\
    &= \mathbb{E} \left( F\left(\boldsymbol{w}_k^G\left(t\right)\right) \right) - \frac{1}{2L} \left\| \nabla F\left(\boldsymbol{w}_k^G\left(t\right)\right) \right\|^2 \\
    & \quad + \frac{1}{L} \mathbb{E} \left(\left(\boldsymbol{o}_k-\boldsymbol{e}_k\right)^T\boldsymbol{n}_k\right) + \frac{1}{2L} \mathbb{E} \left( \left\|\boldsymbol{o}_k - \boldsymbol{e}_k\right\|^2 \right) \\
    & \quad + \frac{1}{2L} \mathbb{E} \left(\left\|\boldsymbol{n}_k\right\|^2\right),
\end{split}
\end{equation}
where (a) stems from the fact that
\begin{equation} \label{eq:(a)}
\begin{aligned}
    &\frac{L \alpha^2}{2} \left\| \nabla F\left(\boldsymbol{w}_k^G\left(t\right)\right) - \boldsymbol{o}_k + \boldsymbol{e}_k + \boldsymbol{n}_k\right\|^2 \\
    &= \frac{1}{2L} \left\| \nabla F\left(\boldsymbol{w}_k^G\left(t\right)\right) \right\|^2 - \frac{1}{L} \left(\boldsymbol{o}_k - \boldsymbol{e}_k + \boldsymbol{n}_k\right)^T \nabla F\left(\boldsymbol{w}_k^G\left(t\right)\right) \\
    &\quad + \frac{1}{2L} \left\| \boldsymbol{o}_k - \boldsymbol{e}_k + \boldsymbol{n}_k\right\|^2.
\end{aligned}
\end{equation}
Next, we first derive $\mathbb{E}\left(\left\|\boldsymbol{o}_k - \boldsymbol{e}_k\right\|^2\right)$, and then compute $\mathbb{E} \left(\left\|\boldsymbol{n}_k\right\|^2\right)$ and $\mathbb{E} \left(\left(\boldsymbol{o}_k-\boldsymbol{e}_k\right)^T\boldsymbol{n}_k\right)$. $\mathbb{E}\left(\left\|\boldsymbol{o}_k - \boldsymbol{e}_k\right\|^2\right)$ can be expressed as 
% \begin{equation} \label{eq:okek}
% \begin{aligned}
% &\mathbb{E}\left(\left\|\boldsymbol{o}_k - \boldsymbol{e}_k\right\|^2\right) \\
% &=\mathbb{E}\left(\left\|\nabla F(\boldsymbol{w}_k^G\left(t\right)) - \frac{\sum\limits_{i \in \mathcal{U}_{k,t}} a_i\left(t\right) \sum\limits_{m=1}^{X_i} \nabla l_{k,i,m}^t}{\sum\limits_{i \in \mathcal{U}_{k,t}}X_ia_i\left(t\right)}\right\|^2\right) \\
% &=\mathbb{E}\left(\left\|\frac{\sum\limits_{i \in \mathcal{C}_1} \sum\limits_{m=1}^{X_i}\nabla l_{k,i,m}^t}{\sum\limits_{i \in \mathcal{T}_k}X_i}\right.\right.\\
% &\left.\left. \quad -\frac{\sum\limits_{i \in \mathcal{T}_k} \!\!X_i \!\sum\limits_{i \in \mathcal{C}_3}\! \sum\limits_{m=1}^{X_i}\!\nabla l_{k,i,m}^t\!-\!\!\sum\limits_{i \in \mathcal{C}_3}\!\! X_i \!\sum\limits_{i \in \mathcal{C}_2} \sum\limits_{m=1}^{X_i}\!\nabla l_{k,i,m}^t}{\sum\limits_{i \in \mathcal{T}_k}X_i\sum\limits_{i \in \mathcal{C}_3}X_i}\right\|^2\right) \\
% &\leq\mathbb{E}\left(\frac{\sum\limits_{i \in \mathcal{C}_1} \sum\limits_{m=1}^{X_i}\|\nabla l_{k,i,m}^t\|}{\sum\limits_{i \in \mathcal{T}_k}X_i}\right.\\
% &\left. \quad +\frac{\sum\limits_{i \in \mathcal{T}_k}\!\! X_i\!\! \sum\limits_{i \in \mathcal{C}_3}\! \sum\limits_{m=1}^{X_i}\!\|\nabla l_{k,i,m}^t\|\!\!-\!\!\!\sum\limits_{i \in \mathcal{C}_3}\! X_i \!\sum\limits_{i \in \mathcal{C}_2}\! \sum\limits_{m=1}^{X_i}\!\|\nabla l_{k,i,m}^t\|}{\sum\limits_{i \in \mathcal{T}_k}X_i\sum\limits_{i \in \mathcal{C}_3}X_i}\right)^2\!\!,
% \end{aligned}
% \end{equation}
\begin{equation} \label{eq:okek_starter}
\begin{aligned}
&\mathbb{E}\left(\left\|\boldsymbol{o}_k - \boldsymbol{e}_k\right\|^2\right) \\
&=\mathbb{E}\left(\left\|\nabla F\left(\boldsymbol{w}_k^G\left(t\right)\right) - \frac{\sum\limits_{i \in \mathcal{U}_{k,t}} a_i\left(t\right) \sum\limits_{m=1}^{X_i} \nabla l_{k,i,m}^t}{\sum\limits_{i \in \mathcal{U}_{k,t}}X_ia_i\left(t\right)}\right\|^2\right)
\end{aligned}
\end{equation}
To further analyze $\mathbb{E}\left(\left\|\boldsymbol{o}_k - \boldsymbol{e}_k\right\|^2\right)$, we rewrite the equation as
\begin{equation} \label{eq:okek}
\begin{aligned}
&\mathbb{E}\left(\left\|\boldsymbol{o}_k - \boldsymbol{e}_k\right\|^2\right) \\
% &=\mathbb{E}\left(\left\|\nabla F\left(\boldsymbol{w}_k^G\left(t\right)\right) - \frac{\sum\limits_{i \in \mathcal{U}_{k,t}} a_i\left(t\right) \sum\limits_{m=1}^{X_i} \nabla l_{k,i,m}^t}{\sum\limits_{i \in \mathcal{U}_{k,t}}X_ia_i\left(t\right)}\right\|^2\right) \\
&=\mathbb{E}\left(\left\|\frac{\sum\limits_{i \in \mathcal{C}_1} \sum\limits_{m=1}^{X_i}\nabla l_{k,i,m}^t}{\sum\limits_{i \in \mathcal{T}_k}X_i}\right.\right.\\
&\left.\left. \quad -\frac{\sum\limits_{i \in \mathcal{T}_k} \!\!X_i \!\sum\limits_{i \in \mathcal{C}_3}\! \sum\limits_{m=1}^{X_i}\!\nabla l_{k,i,m}^t\!-\!\!\sum\limits_{i \in \mathcal{C}_3}\!\! X_i \!\sum\limits_{i \in \mathcal{C}_2} \sum\limits_{m=1}^{X_i}\!\nabla l_{k,i,m}^t}{\sum\limits_{i \in \mathcal{T}_k}X_i\sum\limits_{i \in \mathcal{C}_3}X_i}\right\|^2\right),\\
& \leq\mathbb{E}\left(\frac{\sum\limits_{i \in \mathcal{C}_1} \sum\limits_{m=1}^{X_i}\left\|\nabla l_{k,i,m}^t\right\|}{\sum\limits_{i \in \mathcal{T}_k}X_i}\right.\\
&\left. \quad +\frac{\sum\limits_{i \in \mathcal{T}_k}\!\! X_i\!\! \sum\limits_{i \in \mathcal{C}_3}\! \sum\limits_{m=1}^{X_i}\!\left\|\nabla l_{k,i,m}^t\right\|\!\!-\!\!\!\sum\limits_{i \in \mathcal{C}_3}\! X_i \!\sum\limits_{i \in \mathcal{C}_2}\! \sum\limits_{m=1}^{X_i}\!\left\|\nabla l_{k,i,m}^t\right\|}{\sum\limits_{i \in \mathcal{T}_k}X_i\sum\limits_{i \in \mathcal{C}_3}X_i}\right)^2\!\!,
\end{aligned}
\end{equation}
% where $l_{k,i,m}^t$ is short for $l\left(\boldsymbol{w}_k^G\left(t\right), \boldsymbol{x}_{i,m}, y_{i,m}\right)$,
% $\mathcal{C}_1 = \left\{a_i\left(t\right) = 0 | i \in \mathcal{T}_k\right\}$ is the set of users that belong to $\mathcal{T}_k$ but do not join CFL training of task $k$, $\mathcal{C}_2 = \left\{a_i\left(t\right) = 1 | i \in \mathcal{T}_k\right\}$ is the set of users that belong to $\mathcal{T}_k$, and $\mathcal{C}_3 = \left\{a_i\left(t\right) = 1 | i \in \mathcal{U}_{k,t}\right\}$ is the set of users that belong to other task user sets $\left\{k'\neq k | k' \in \mathcal{K}\right\}$ and join CFL training of task $k$ due to incorrect identification. Then, we have $\mathcal{C}_1 \cup \mathcal{C}_2 = \mathcal{T}_k$. Since $....$ \cite{chen2020joint}, $\mathbb{E}\left(\left\|\boldsymbol{o}_k - \boldsymbol{e}_k\right\|^2\right)$ can be rewritten as
% \begin{equation} \label{eq:okek}
% \begin{aligned}
% &\mathbb{E}\left(\left\|\boldsymbol{o}_k - \boldsymbol{e}_k\right\|^2\right) 
% \leq\mathbb{E}\left(\frac{\sum\limits_{i \in \mathcal{C}_1} \sum\limits_{m=1}^{X_i}\left\|\nabla l_{k,i,m}^t\right\|}{\sum\limits_{i \in \mathcal{T}_k}X_i}\right.\\
% &\left. \quad +\frac{\sum\limits_{i \in \mathcal{T}_k}\!\! X_i\!\! \sum\limits_{i \in \mathcal{C}_3}\! \sum\limits_{m=1}^{X_i}\!\left\|\nabla l_{k,i,m}^t\right\|\!\!-\!\!\!\sum\limits_{i \in \mathcal{C}_3}\! X_i \!\sum\limits_{i \in \mathcal{C}_2}\! \sum\limits_{m=1}^{X_i}\!\left\|\nabla l_{k,i,m}^t\right\|}{\sum\limits_{i \in \mathcal{T}_k}X_i\sum\limits_{i \in \mathcal{C}_3}X_i}\right)^2\!\!,
% \end{aligned}
% \end{equation}
where $l_{k,i,m}^t$ is short for $l\left(\boldsymbol{w}_k^G\left(t\right), \boldsymbol{x}_{i,m}, y_{i,m}\right)$,
$\mathcal{C}_1 = \left\{a_i\left(t\right) = 0 | i \in \mathcal{T}_k\right\}$ is the set of users that belong to $\mathcal{T}_k$ but do not join CFL training of task $k$, $\mathcal{C}_2 = \left\{a_i\left(t\right) = 1 | i \in \mathcal{T}_k\right\}$ is the set of users that belong to $\mathcal{T}_k$, and $\mathcal{C}_3 = \left\{a_i\left(t\right) = 1 | i \in \mathcal{U}_{k,t}\right\}$ is the set of users that belong to other task user sets $\left\{k'\neq k | k' \in \mathcal{K}\right\}$ and join CFL training of task $k$ due to incorrect identification. Then, we have $\mathcal{C}_1 \cup \mathcal{C}_2 = \mathcal{T}_k$. The inequality in (\ref{eq:okek}) stems from triangle-inequality. 
Based on the assumption in (19) (i.e., $ \left\|\nabla l_{k,i,m}^t\right\| \leq \sqrt{\zeta_1 + \zeta_2 \left\|\nabla F\left(\boldsymbol{w}_{k}^G\left(t\right)\right)\right\|^2}$), $\mathbb{E}\left(\left\|\boldsymbol{o}_k-\boldsymbol{e}_k\right\|^2\right)$ can be expressed by
% \begin{equation}
% \begin{aligned} \label{eq:sim_okek}
%     &\mathbb{E}(\|\boldsymbol{o}_k-\boldsymbol{e}_k\|^2) \\
%     &\leq \mathbb{E}\left(\frac{\sum\limits_{i \in \mathcal{C}_1}\!X_i\! +\!\! \sum\limits_{i \in \mathcal{T}_k}\!X_i\! -\!\! \sum\limits_{i \in \mathcal{C}_2}\!X_i}{\sum\limits_{i \in \mathcal{T}_k}X_i}\right)\!\!\times\!(\zeta_1 \!+ \zeta_2 \|\nabla F(\boldsymbol{w}_{k}^G\left(t\right))\|^2) \\
%     & = 4\frac{\mathbb{E}\left(\sum\limits_{i \in \mathcal{C}_1}X_i\right)^2}{\left(\sum\limits_{i \in \mathcal{T}_k}X_i\right)^2} (\zeta_1 + \zeta_2 \|\nabla F(\boldsymbol{w}_{k}^G\left(t\right))\|^2)\\
%     &\leq 4\left(1 - \frac{\mathbb{E}\left(\sum\limits_{i \in \mathcal{C}_2}X_i\right)}{\sum\limits_{i \in \mathcal{T}_k}X_i}\right)(\zeta_1 + \zeta_2 \|\nabla F(\boldsymbol{w}_{k}^G\left(t\right))\|^2).
% \end{aligned}
% \end{equation}
\begin{equation} \label{eq:sim_okek1}
\begin{aligned}
    &\mathbb{E}\left(\left\|\boldsymbol{o}_k-\boldsymbol{e}_k\right\|^2\right) \\
    &\leq \mathbb{E}\left(\!\!\frac{\sum\limits_{i \in \mathcal{C}_1}\!X_i\! +\!\! \sum\limits_{i \in \mathcal{T}_k}\!X_i\! -\!\! \sum\limits_{i \in \mathcal{C}_2}\!X_i}{\sum\limits_{i \in \mathcal{T}_k}X_i}\!\!\right)\!\!\!\left(\!\zeta_1 \!+ \zeta_2 \left\|\nabla F\left(\boldsymbol{w}_{k}^G\left(t\right)\right)\right\|^2\!\right)\\
    &= 4\frac{\mathbb{E}\left(\sum\limits_{i \in \mathcal{C}_1}X_i\right)^2}{\left(\sum\limits_{i \in \mathcal{T}_k}X_i\right)^2} \left(\zeta_1 + \zeta_2 \left\|\nabla F\left(\boldsymbol{w}_{k}^G\left(t\right)\right)\right\|^2\right).
\end{aligned}
\end{equation}
Since $\sum\limits_{i \in \mathcal{C}_1}X_i \leq \sum\limits_{i \in \mathcal{T}_k}X_i$, (\ref{eq:sim_okek1}) can be simplified as
\begin{equation} \label{eq:sim_okek}
\begin{aligned}
    \mathbb{E}\left(\left\|\boldsymbol{o}_k-\boldsymbol{e}_k\right\|^2\right) & \leq 4\left(1 - \frac{\mathbb{E}\left(\sum\limits_{i \in \mathcal{C}_2}X_i\right)}{\sum\limits_{i \in \mathcal{T}_k}X_i}\right) \\
    & \quad \times \left(\zeta_1 + \zeta_2 \left\|\nabla F\left(\boldsymbol{w}_{k}^G\left(t\right)\right)\right\|^2\right).
\end{aligned}
\end{equation}
Since $\mathbb{E}\left(\sum\limits_{i \in \mathcal{C}_2}X_i\right) = \sum\limits_{i \in \mathcal{T}_k}X_i\left(1-p_{k,i}^t\right)a_i\left(t\right)$, we have
% \begin{equation} \label{eq:ec2x}
% \begin{aligned}
%     \mathbb{E}\left(\sum\limits_{i \in \mathcal{C}_2}X_i\right) = \sum\limits_{i \in \mathcal{T}_k}X_i(1-p_{k,i}^t)a_i\left(t\right)),
% \end{aligned}
% \end{equation}
% where $p_a^k\left(t\right)$ is the probability of each user classified as task $k$ and connected to BSs.
% In (\ref{eq:ec2x}), $p_a^k\left(t\right)$ can be expressed by
% \begin{equation}
% \begin{aligned} \label{eq:pka}
%     p_a^k\left(t\right) = \frac{\mathbb{E}(|\mathcal{C}_3)|}{\mathbb{E}(|\mathcal{U}_{k,t})|},
% \end{aligned}
% \end{equation}
% where
% \begin{equation} \label{eq:num_uk}
%     \mathbb{E}(|\mathcal{U}_{k,t}|) = \sum\limits_{i \in \mathcal{T}_k}(1-p_{k,i}^t) + \frac{1}{K-1}\sum\limits_{k'\neq k}\sum\limits_{i \in \mathcal{T}_{k'}}p_{k',i}^t.
% \end{equation}
% In (\ref{eq:pka}), \( p_a^k\left(t\right) \) is derived by taking the expectation of the number of selected users in \( \mathcal{U}_{k,t} \) divided by the expectation of the total number of users in \( \mathcal{U}_{k,t} \), representing the user association probability within \( \mathcal{U}_{k,t} \).
% Given (\ref{eq:ec2x}), (\ref{eq:sim_okek}) can be simplified as follows:
\begin{equation} \label{eq:final_okek}
\begin{aligned}
    &\mathbb{E} \left(\left\|\boldsymbol{o}_k - \boldsymbol{e}_k\right\|^2 \right) \\
    &\leq 4 \!\! \left(\! 1 - \frac{\sum\limits_{i \in \mathcal{T}_k} X_i \left( 1 \!-\! p_{k,i}^t \right) a_i\left(t\right)}{\sum\limits_{i \in \mathcal{T}_k} X_i} \!\right) \!\!
    \left(\! \zeta_1 \!+\! \zeta_2 \left\|\!\nabla\! F\!\left(\boldsymbol{w}_{k}^G\left(t\right)\right)\!\right\|^2 \!\right) \\
    &= 4 Q_{0,k,t} \left( \zeta_1\! +\! \zeta_2 \left\|\nabla F\left(\boldsymbol{w}_{k}^G\left(t\right)\right)\right\|^2 \right),
\end{aligned}
\end{equation}
where $Q_{0,k,t} = 1 - \frac{1}{\sum\limits_{i \in \mathcal{T}_k}X_i}\sum\limits_{i \in \mathcal{T}_k}X_i\left(1-p_{k,i}^t\right)a_i\left(t\right)$. 

Next, we calculate $\mathbb{E} \left(\left\|\boldsymbol{n}_k\right\|^2\right)$ and $\mathbb{E} \left(\left(\boldsymbol{o}_k-\boldsymbol{e}_k\right)^T\boldsymbol{n}_k\right)$ in~(\ref{eq:transfer}).
Since the DP noise is independent of user connection variables $a_i\left(t\right)$ and $\boldsymbol{n}_{i}\left(t\right) = \left(n_{i,e}\left(t\right)\right)_{e \in [d]} \sim N\left(0,\sigma^2_{i}\mathbf{I}_d\right)$, we have
\begin{equation}\label{eq:eoen}
\begin{split}
    \mathbb{E} \left(\left(\boldsymbol{o}_k - \boldsymbol{e}_k\right)^T \boldsymbol{n}_k\right) &= \mathbb{E} \left(\left(\boldsymbol{o}_k - \boldsymbol{e}_k\right)^T\right) \mathbb{E} \left(\boldsymbol{n}_k\right) = 0,
\end{split}
\end{equation}
%Next, we calculate $\mathbb{E} \left(\left\|\boldsymbol{n}_k\right\|^2\right)$ as follows:
\begin{equation}\label{eq:ee}
\begin{aligned}
    &\mathbb{E} \left(\left\|\boldsymbol{n}_k\right\|^2\right) = \mathbb{E}\left\|\frac{1}{\sum\limits_{i \in \mathcal{U}_{k,t}}X_ia_i\left(t\right)}\sum\limits_{i \in \mathcal{U}_{k,t}}X_ia_i\left(t\right)\boldsymbol{n}_i\left(t\right)\right\|^2 \\
    &= \mathbb{E} \left\|\frac{1}{\sum\limits_{i \in \mathcal{U}_{k,t}}X_ia_i\left(t\right)}\sum\limits_{i \in \mathcal{U}_{k,t}}X_ia_i\left(t\right)\sum\limits_{e=1}^d n_{i,e}\left(t\right)\right\|^2 \\
    &\leq \mathbb{E} \left\|\frac{1}{\sum\limits_{i \in \mathcal{U}_{k,t}}X_ia_i\left(t\right)}\sum\limits_{i \in \mathcal{U}_{k,t}}X_ia_i\left(t\right)\sum\limits_{e=1}^d \sigma_{max}\right\|^2 \\
    &= d^2\sigma_{max}^2,
\end{aligned}
\end{equation}
Substituting (\ref{eq:final_okek})-(\ref{eq:ee}) into (\ref{eq:transfer}), we have
\begin{equation} \label{eq:ad_F}
\begin{aligned}
    &\mathbb{E} \left( F\left(\boldsymbol{w}_k^G\left(t+1\right)\right) \right) \leq \mathbb{E} \left( F\left(\boldsymbol{w}_k^G\left(t\right)\right) \right) \\
    &\quad + \frac{2 \zeta_1 }{L}Q_{0,k,t}\! - \!\frac{1-4\zeta_2}{2L}Q_{0,k,t} \left\|\!\nabla F\left(\boldsymbol{w}_{k}^G\!\left(t\right)\right)\!\right\|^2 \!+\! \frac{d^2\sigma_{max}^2}{2L},
\end{aligned}
\end{equation} 
Subtract $\mathbb{E}\left(F\left(\boldsymbol{w}_k^*\right)\right)$ from both sides of (\ref{eq:ad_F}), we have
\begin{equation} \label{eq:fi_F}
\begin{aligned}
    &\mathbb{E} \left( F\left(\boldsymbol{w}_k^G\left(t+1\right)\right) - F\left(\boldsymbol{w}_k^*\right) \right)\\ 
    &\leq \mathbb{E} \left( F\left(\boldsymbol{w}_k^G\left(t\right)\right) - F\left(\boldsymbol{w}_k^*\right) \right)+ \frac{2 \zeta_1 }{L}Q_{0,k,t} \\
    & \quad \!-\! \frac{1-4\zeta_2}{2L}Q_{0,k,t} \left\|\nabla F\left(\boldsymbol{w}_{k}^G\left(t\right)\right)\right\|^2\! + \!\frac{d^2\sigma_{max}^2}{2L}.
\end{aligned}
\end{equation}
Given (17) and (18), we have
\begin{equation} \label{eq:noidea}
    \left\|\nabla F\left(\boldsymbol{w}_k^G\left(t\right)\right)\right\|^2 \geq 2\mu\left(F\left(\boldsymbol{w}_k^G\left(t\right)\right) - F\left(\boldsymbol{w}_k^*\right)\right).
\end{equation}
Substituting (\ref{eq:noidea}) into (\ref{eq:fi_F}), we have
\begin{equation} \label{eq:one_time}
\begin{aligned}
    &\mathbb{E} \left( F\left(\boldsymbol{w}_k^G\left(t+1\right)\right) - F\left(\boldsymbol{w}_k^*\right) \right) \\
    &\leq \frac{2\zeta_1Q_{1,k,t}}{L- \mu + 4 \mu \zeta_2} \!+ \!\frac{d^2\sigma_{max}^2}{2L} \!+ \!Q_{1,k,t}\!\left(F\!\left(\boldsymbol{w}_k^G\left(t\right)\right) \!-\! F\!\left(\!\boldsymbol{w}_k^*\!\right)\right)\!,
\end{aligned}
\end{equation}
where $Q_{1,k,t} = \frac{L- \mu + 4 \mu \zeta_2}{L}Q_{0,k,t}$. 
% Applying (\ref{eq:one_time}) recursively, we have
% \begin{equation} \label{eq:multiple}
% \begin{aligned}
%     &\mathbb{E} \left( F(\boldsymbol{w}_k^G(t+1)) - F(\boldsymbol{w}_k^*) \right) \\
%     &\leq \frac{2\zeta_1}{L- \mu + 4 \mu \zeta_2}\sum\limits_{k=0}^{t-1}Q_2^{k+1}
%     + Q_2^t \mathbb{E} \left( F(\boldsymbol{w}_k^G(0)) - F(\boldsymbol{w}_k^*) \right) \\
%     & \quad + \frac{Q_1}{2L}\sum\limits_{k=0}^{t-1}Q_2^{k+1}\\
%     & =\frac{2\zeta_1}{L- \mu + 4 \mu \zeta_2}\frac{Q_2-Q_2^{t+1}}{1-Q_2} + \frac{Q_1(1-Q_2^t)}{2L(1-Q_2)}\\
%     & \quad + Q_2^t \mathbb{E} \left( F(\boldsymbol{w}_k^G(0)) - F(\boldsymbol{w}_k^*) \right)
% \end{aligned}
% \end{equation}
This completes the proof.
\hfill $\Box$

\bibliography{ref}

\bibliographystyle{IEEEtran}

\end{document}